\documentclass[review]{fcs}
\usepackage{cite}
\usepackage{amsmath,amssymb,amsfonts}
\usepackage{algorithmic}
\usepackage{algorithm}
\usepackage{graphicx}
\usepackage{subfigure}
\usepackage{textcomp}
\usepackage{xcolor}
\usepackage{makecell}
\usepackage{multirow}
\usepackage{mwe} 

\def\BibTeX{{\rm B\kern-.05em{\sc i\kern-.025em b}\kern-.08em
    T\kern-.1667em\lower.7ex\hbox{E}\kern-.125emX}}
\title{HyperTuner: A Cross-Layer Multi-Objective Hyperparameter Auto-Tuning Framework for Data Analytic Services \\}
\author[1]{Hui Dou}\author[1]{Shanshan Zhu}\author*[1]{Yiwen Zhang}\author[2]{Pengfei Chen}\author[2]{Zibin Zheng}
\address[1]{ School of Computer Science and Technology, Anhui University, Hefei 230601, China}
\address[2]{School of Computer Science and Engineering, Sun Yat-Sen University, Guangzhou 510006, China}
\fcssetup{
  received       = {month dd, yyyy},
  accepted       = {month dd, yyyy},
  corr-email     = {zhangyiwen@ahu.edu.cn},
}
\begin{abstract}
Hyper-parameters optimization (HPO) is vital for machine learning models. Besides model accuracy, other tuning intentions such as model training time and energy consumption are also worthy of attention from data analytic service providers. Hence, it is essential to take both model hyperparameters and system parameters into consideration to execute cross-layer multi-objective hyperparameter auto-tuning. Towards this challenging target, we propose HyperTuner in this paper. To address the formulated high-dimensional black-box multi-objective optimization problem, HyperTuner first conducts multi-objective parameter importance ranking with its MOPIR algorithm and then leverages the proposed ADUMBO algorithm to find the Pareto-optimal configuration set. During each iteration, ADUMBO selects the most promising configuration  from the generated Pareto candidate set via maximizing a new well-designed metric, which can adaptively leverage the uncertainty as well as the predicted mean across all the surrogate models along with the iteration times. We evaluate HyperTuner on our local distributed TensorFlow cluster and experimental results show that it is always able to find a better Pareto configuration front superior in both convergence and diversity compared with the other four baseline algorithms. Besides, experiments with different training datasets, different optimization objectives and different machine learning platforms verify that HyperTuner can well adapt to various data analytic service scenarios.
\end{abstract}


\begin{document}
\begin{sloppypar}
\section{Introduction}
\label{sec:1}
Machine learning especially deep learning models now play an irreplaceable role to delivery various data analytic services such as image recognition\cite{pouyanfar2018survey} and anomaly detection\cite{pang2021deep}.  Since this sort of services are often recurrently invoked with different datasets, tuning hyperparameters for their internal models becomes an essential prerequisite course before service deployment. However, considering the remarkable time consumption, resource occupation and energy consumption to evaluate each candidate hyperparameter configuration, there are an urgent demand from data analytic service providers for efficiently automatic hyperparameter optimization (HPO). In the past few years, many enlightening literatures from both academia and industry have been published and step closer towards this target\cite{kotthoff2019auto,li2017hyperband,falkner2018bohb,akiba2019optuna}.

\begin{figure*}
 \centering
 \subfigure{    
  \label{fig:1a}\includegraphics[width=0.32\textwidth]{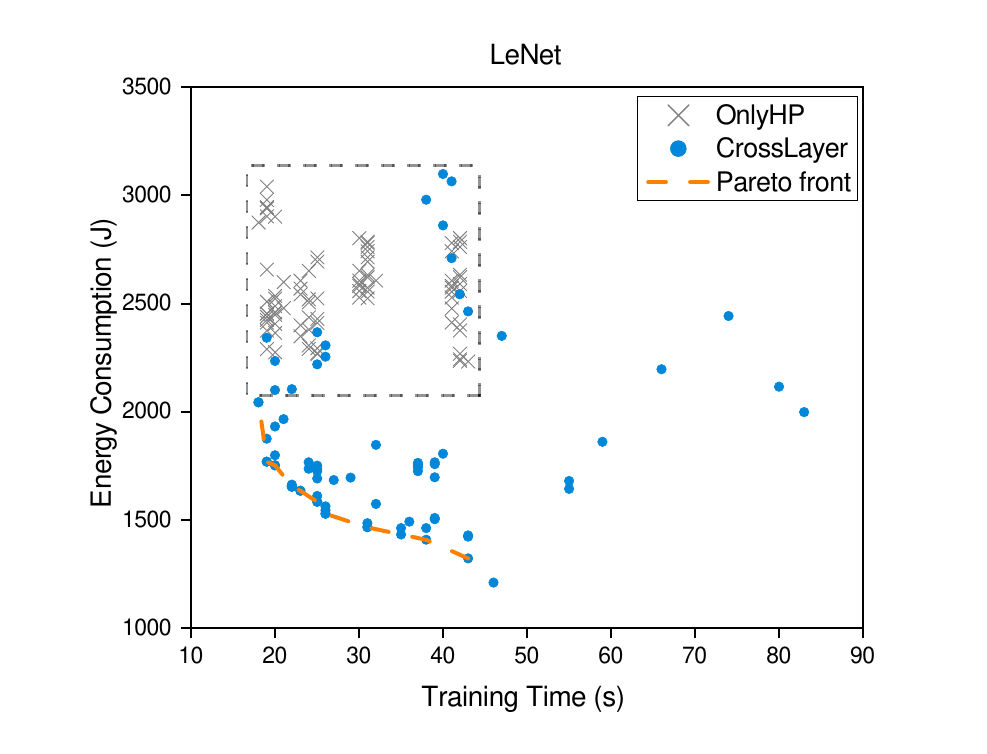}}
 \subfigure{  
  \label{fig:1b}\includegraphics[width=0.32\textwidth]{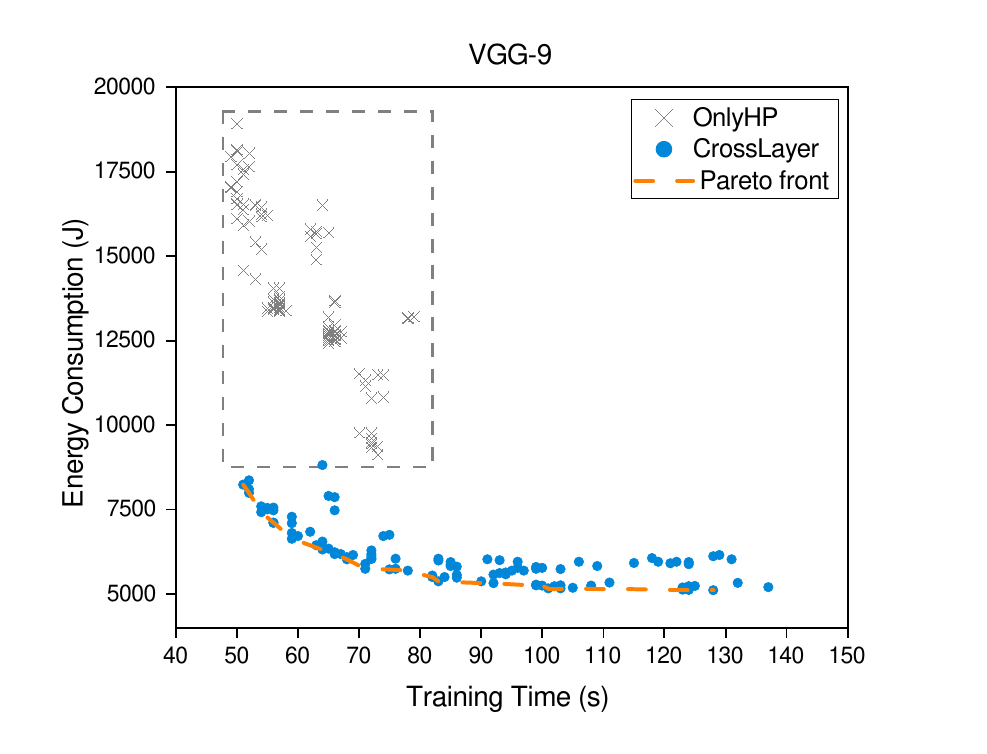}}
 \subfigure{
  \label{fig:1c}\includegraphics[width=0.32\textwidth]{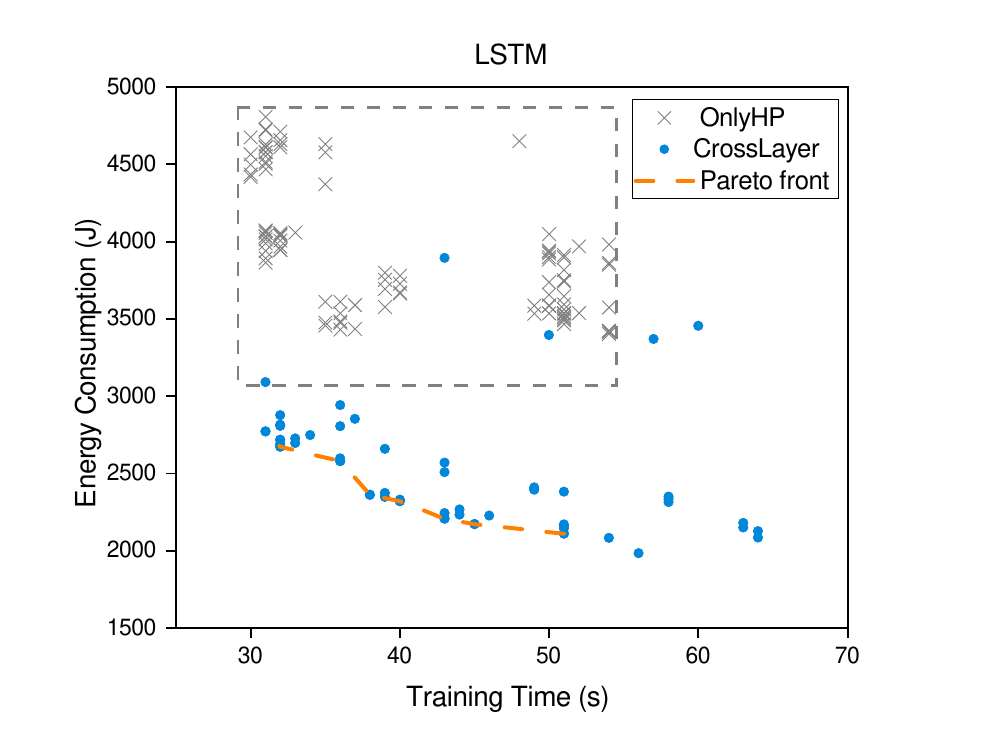}}
 \caption{Necessity of considering system parameters when executing multi-objective HPO for LeNet, VGG-9 and LSTM model training tasks.}
 \label{fig:1}
\end{figure*}

Unfortunately, conventional HPO studies often solely focus on the model accuracy and totally ignore the associated time consumption, energy consumption or carbon emission during the model training procedure, which are of equal importance for service providers especially considering the 
frequently repeated model training tasks. As shown in Fig.\ref{fig:1a}, the training time of a LeNet model to achieve the same accuracy level can vary from 20 to 45 seconds with different hyperparameter configurations (\emph{marked as OnlyHP}) on our local TensorFlow cluster. Besides, training BERT on GPU generates roughly equivalent carbon footprint to a trans-American flight\cite{strubell2019energy}. Therefore, it is necessary for data analytic service providers to conduct multi-objective hyperparameter optimization. 

Recently, several representative work from different scenarios have already tried to solve the multi-objective HPO problem\cite{morales2022survey, ann, pal,hernandez2016predictive,ehi,sms-ego }. However, we find that most of them overlooked the truth that many optimization objectives are not only related to the model hyperparameter settings, but also closely bounded up with parameters from the whole machine learning system stack. For example, besides the hyperparameters, the model training time is also highly dependent of the parameters from the layer of operating system kernel as well as the machine learning platform. We denote them as system parameters in this paper to distinguish with the model hyperpaprameters. Specifically, in order to verify the necessity of system parameters when executing multi-objective HPO for data analytic services, we trained three different deep learning models respectively with grid-search hyperparameter configurations (\emph{marked as OnlyHP}) as well as 60 randomly generated $<$hyperparameter, system parameter$>$ configurations (\emph{marked as CrossLayer}) on our local distributed TensorFlow cluster. It is worth noting that the training time and the energy consumption of CPU cores are two optimization objectives and the model accuracy should be higher than the pre-specified constraint in this toy experiment. Fig.\ref{fig:1} illustrates that integrating HPO with system parameter tuning is able to find a much better Pareto front for all the three deep learning models. Hence, data analytic service providers should simultaneously take the cross-layer system parameters and model hyperparameters into account to tackle with their multiple optimization objectives in practice.

In summary, there are three major challenges to achieve this target for practical data analytic service scenarios: First of all, the cross-layer system parameters and model hyperparameters related to different optimization objectives together form a high dimensional configuration space; Second, the configuration space is maddeningly complex since each parameter can be a variable of enumeration, integer and even floating point; Third, the evaluation cost of each candidate configuration cannot be directly overlooked. Simultaneously tackling with these challenges is necessary for multi-objective HPO but not an easy thing. For instance, as the most related work, FlexiBO\cite{iqbal2020flexibo} also tries to tune cross-layer parameters for multiple objectives based on the Pareto Active Learning algorithm\cite{pal}. However, its effectiveness and efficiency highly depends on the assumption of a discrete candidate configuration space since it has to traverse the whole space to generate the next sample configuration. In this paper, we propose HyperTuner, a cross-layer multi-objective hyperparameter auto-tuning framework for data analytic services. Specifically, to simultaneously tackle with above three challenges, HyperTuner combines the following two novel techniques: \textbf{1) MOPIR (Parameter Importance Ranking for Multiple Objectives):} To tackle with the challenge of high-dimensional configuration space, we design MOPIR to conduct parameter importance ranking for multiple objectives. The core idea of MOPIR is to obtain the Pareto importance rank of each parameter based on their Gini index along each objective as well as the non-dominated sorting method. Detailed description of MOPIR
can be found in Sec. \ref{sec:4a}. \textbf{2) ADUMBO (Multi-Objective Bayesian Optimization with ADaptive Uncertainty):} Considering the complex configuration space and high configuration evaluation cost, we also propose ADUMBO to solve the corresponding black-box multi-objective optimization problem. Through introducing a novel adaptive uncertainty metric to guide the
choice of next promising configuration to be evaluated during each iteration, ADUMBO further improves the sample-efficiency feature of vanilla Bayesian optimization for multiple objectives. Besides, ADUMBO itself is also a computationally-efficient algorithm compared with prior multi-objective methods. In fact, to generate the next most promising configuration for each iteration, only a cheap mutli-objective optimization problem need to be solved in ADUMBO instead of traversing the whole candidate configuration space like FlexiBO did. Detailed description of ADUMBO can be found in Sec. \ref{sec4b}. 

To evaluate the effectiveness of HyperTuner for auto-tuning cross-layer parameters for data analytic services, we conducted a series of experiments on our local distributed TensorFlow cluster with 3 DNN model training tasks. Overall, compared with other four baseline multi-objective black-box optimization algorithms, HyperTuner can always find a better Pareto-optimal configuration set in both convergence (minimizing the gap between candidate solutions and the true Pareto-optimal set) and diversity (maximizing the distribution of candidate solutions)\cite{metric-moo1,metric-moo2}. In addition, to evaluate the adaptability of HyperTuner, we also conducted additional experiments with different training datasets, different optimization objectives and different machine learning platforms. Experimental results verify that HyperTuner can well adapt to the various application scenarios of data analytic services. The key contributions of this paper are concluded as follows: 
\begin{itemize}
\item With theoretical analysis and experimental verification, we point out that multi-objective hyperparameter optimization should take both model hyperparameters and system parameters into consideration. 
\item To address the challenges associating with the cross-layer multi-objective hyperparameter 
optimization problem, we propose MOPIR to shrink the configuration space and then design ADUMBO to efficiently solve the corresponding black-box multi-objective optimization problem via introducing a novel adaptive uncertainty metric. 
\item Based on above two novel algorithms,we implement a cross-layer multi-objective hyperparameter auto-tuning framework HyperTuner. Extensive experiments based on our local TensorFlow and Spark-based distributed machine learning cluster verify the effectiveness and adaptability of HyperTuner.
Source codes of HyperTuner are available at \url{https://github.com/zss233-21/HyperTuner.git}
\end{itemize}

The rest of this paper is organized as follows. Section \ref{sec:2} introduces the background and gives the problem statement. Section \ref{sec:3} describes the system overview of HyperTuner. Section \ref{sec:4} proposes the two key algorithms and then describes the implementation of HyperTuner. After that, Section \ref{sec:5} introduces our experimental setups. Section \ref{sec:6} presents the experimental results and analysis. Section \ref{sec:7} describes the related work and Section \ref{sec:8} gives the conclusion and future work.
\section{Background and Problem Statement}
\label{sec:2}
\subsection{Cross-Layer Configuration Parameters}
\label{sec:2a}

\begin{figure}[h]
    \centering
    \includegraphics[width=9cm,height=9cm]{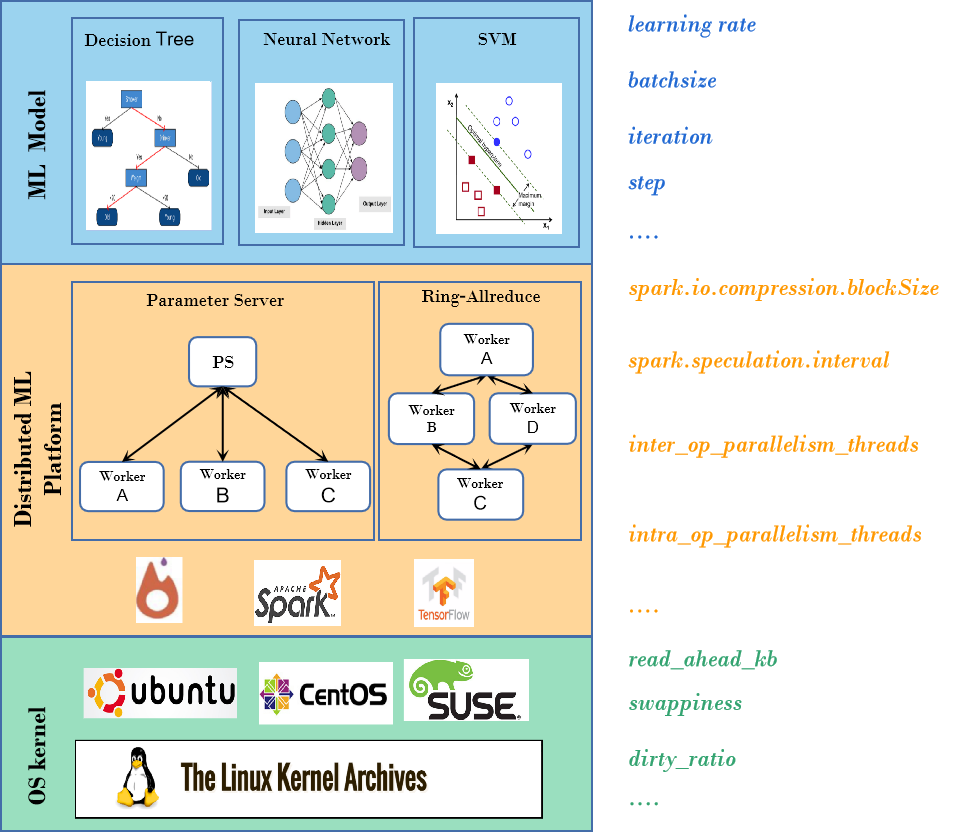}
    \caption{Machine learning system stack and the cross-layer configuration parameters for data analytic services.}
    \label{fig:2}
\end{figure}

\begin{table*}[htbp]
\caption{The five power governors in Linux CPUFreq subsystem.}
\begin{center}
\scalebox{1}{
    \begin{tabular}{|c|c|c|c|}
    \hline
    \textbf{CPUFreq Governor} &\textbf{Short Description}&\textbf{Representative Parameters} \\
    \hline
    performance & Statically sets the CPU frequency to the highest available frequency & \emph{scaling\_max\_freq}	\\
    \hline
    powersave & Statically sets the CPU frequency to the lowest available frequency & \emph{scaling\_min\_freq} \\
    \hline
    ondemand & \makecell[l]{Dynamically adjusts the CPU frequency to the highest available \\frequency when the load is high and gradually degrades the CPU \\frequency when the load is low}& \emph{sampleing\_rate}, \emph{up\_thosehold}\\
    \hline
    conservative & \makecell[l]{Similar to ondemand, the difference is that the frequency\\ increases gradually} & \emph{freq\_step}, \emph{down\_thosehold} \\
    \hline
    userspace & \makecell[l]{Statically sets the CPU frequency to a user-defined value} & $scaling\_setspeed$ \\
    \hline
    \end{tabular}}
\label{cpu}
\end{center}
\end{table*}
Conventional hyperparameter optimization methods usually target on model accuracy only. However, other metrics such as training time and energy consumption are also important for data analytic service providers in practice. In order to achieve multi-objective HPO, we also insist that both model hyperparameters and system parameters should be taken into consideration. Figure \ref{fig:2} shows the underlying machine learning system stack of data analytic services and lists the cross-layer parameters related to different optimization objectives. Taking training time consumption for example, model hyperparameters \emph{learning\_rate} and \emph{batchsize}, system parameters \emph{inter\_op\_parallelism\_threads} and \emph{intra\_
op\_parallelism\_threads} from TensorFlow (especially for the CPU backends), system parameters provided by the CPU frequency scaling subsystem from Linux kernel are all strongly related. In detail, \emph{inter\_op\_pa-
rallelism\_threads} sets the number of threads used for parallelism between independent operations while \emph{intra\_op\_parallelism\_threads} specifies the number of threads used within an individual operation for parallelism, they can greatly affect the performance of TensorFlow workloads runinig on CPUs\cite{tf}. Similarly, tuning CPU frequency settings can directly change the clock frequency and voltage configurations of processors and thus also affects the final model training time\cite{spantidi2020frequency}. More detail discussion about the CPU frequency scaling subsystem can be found in the next immediate subsection. 
\begin{figure*}
 \centering
 \subfigure{    
  \label{fig:3a}\includegraphics[width=0.32\textwidth]{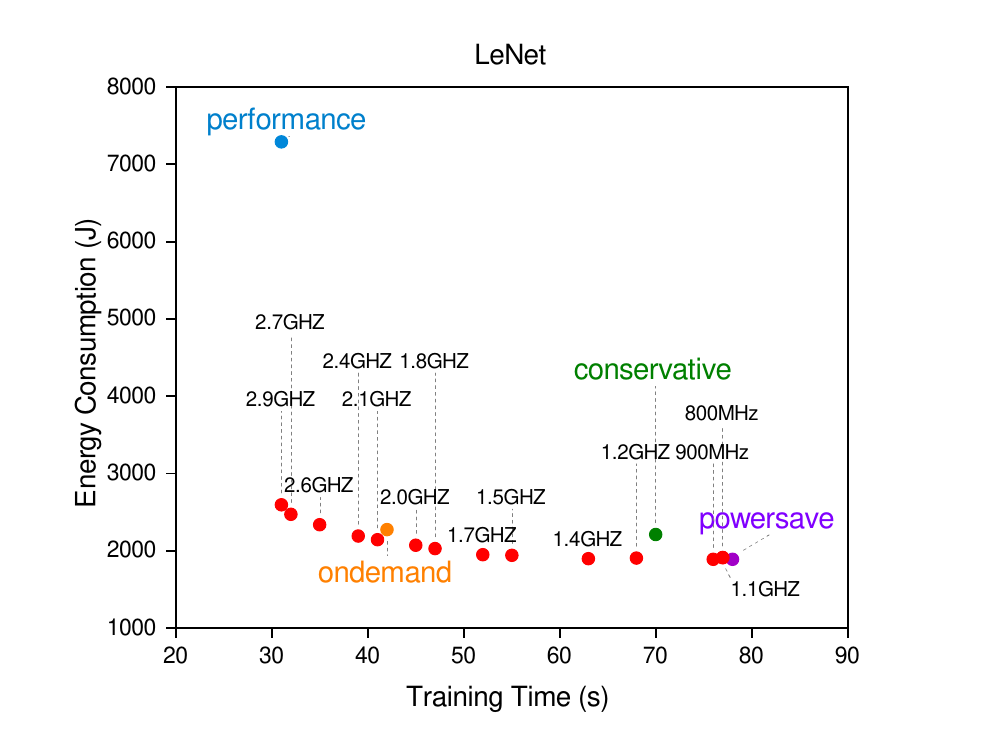}}
 \subfigure{  
  \label{fig:3b}\includegraphics[width=0.32\textwidth]{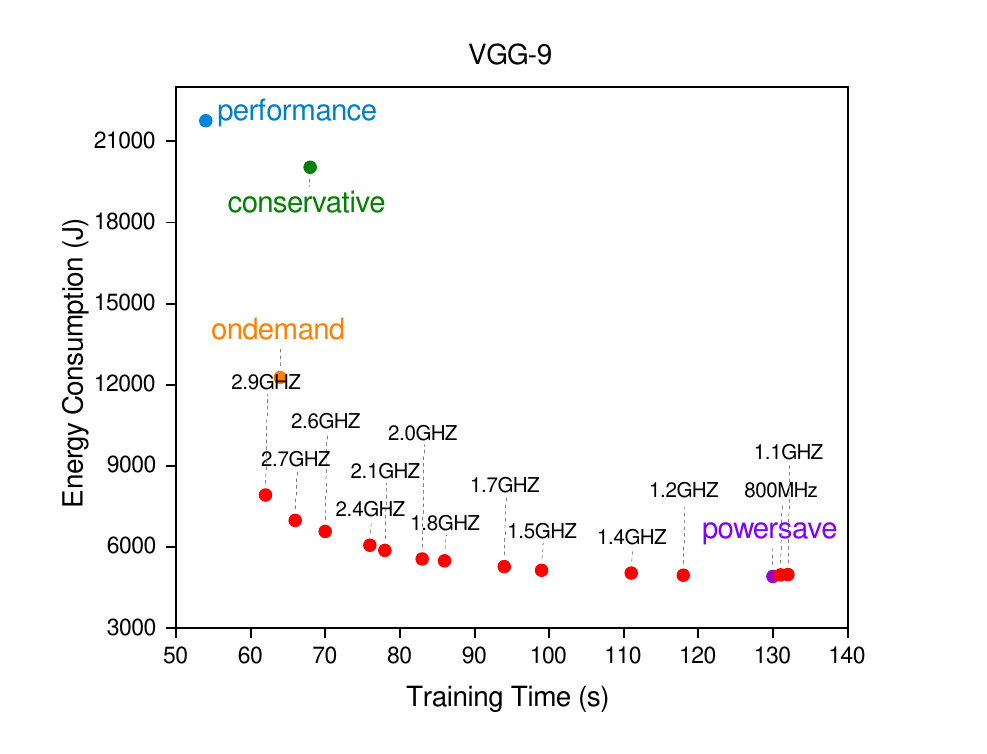}}
 \subfigure{
  \label{fig:3c}\includegraphics[width=0.32\textwidth]{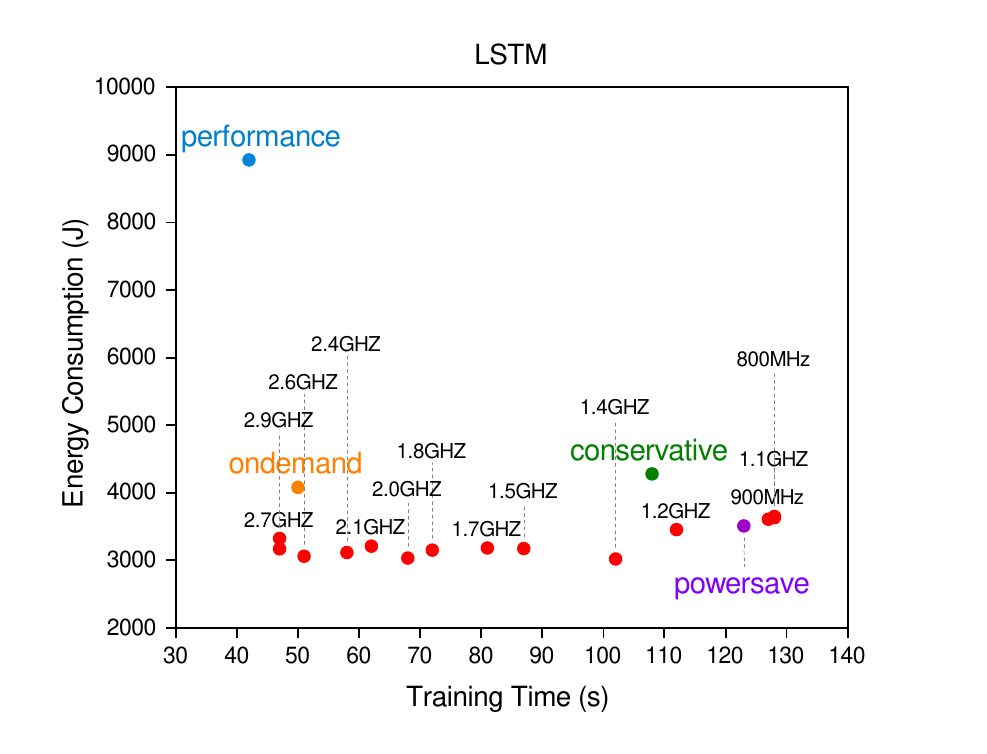}}
 \caption{The performance of five CPUFreq governors when regarding  model training time and the associated energy consumption as  two objectives. }
 \label{fig:governor}
\end{figure*}
\subsection{CPU Frequency Scaling Subsystem and Its Parameters}
\label{sec:2b}
Recently, the energy consumption and carbon emission associated with the model training process have already drawn attentions from both academia and industry\cite{dvfs,strubell2019energy,stamoulis2018hyperpower,energy2}. In fact, to tradeoff between CPU capacity and CPU power drawn, modern processors often provide hardware interfaces to switch across different clock frequency and voltage configurations (i.e., P-states). Based on these hardware interfaces, the Linux kernel implement the CPUFreq (CPU Frequency scaling) subsystem, which includes five different governors since the 2.6.10 version\cite{Linux}. Table \ref{cpu} briefly describes these governors and  also lists several representative parameters for each. Specifically, the \emph{powersave} and \emph{performance} governor statically sets the CPU to its lowest and highest frequency within the borders of parameter \emph{scaling\_min\_freq} and \emph{scaling\_max\_freq}, which fails to adapt to the system load changes. On the contrary, \emph{ondemand} and \emph{conservative}  are two governors which can dynamically adjust the CPU frequency according to the system states. For example, the \emph{ondemand} governor uses current system load as the metric to automatically select CPU frequency for users. In detail, the system load is estimated according to the CPU usage statistics over the last period and parameter \emph{sampling\_rate} determines how often the kernel looks at the CPU usage. However, when the system frequently switches between idle and heavy load, they may lead to low performance/energy gains. Different from all the other governors, the \emph{userspace} governor allows users to directly adjust their desired core frequency from the available frequency set and thus can meet a wide range of system loads. As shown in Figure \ref{fig:governor}, the \emph{userspace} governor is always able to help data analytic service providers find a Pareto configuration set with both better convergence and diversity towards the two optimization objectives compared with other four governors. Therefore, we choose to utilize the \emph{userspace} governor in our experiments and leverage the \emph{cpufrequtils} package to set different CPU frequencies. Noting that the CPU Turbo Boost capacity is always disabled to mitigate the potential uncertainty of experimental results. 
\subsection{Problem Statement}
In this paper, we focus on how to achieve multi-objective hyperparameter optimization for data analytic service providers when taking the cross-layer configuration parameters described above into consideration. Specifically, we assume that there are $D$ configurable parameters in total across the machine learning system stack and use $\boldsymbol{x}$ to denote a candidate configuration, where $x_i$ means the $i$-th ($i=1,2,...,D$) parameter. In practice, $x_i$ can be either numeric or categorical. Noting that for a categorical parameter, we introduce an index to represent its category and set the index to be the value of this parameter. Therefore, the candidate configuration space can be expressed as $\mathbb{X}$ =$\prod_{i=1}^{D} ({x_i})$. Besides, we also assume the service provider cares about $M$ different optimization objectives and use $f_m$ to denote the $m$-th ($m=1,2,...,M$) objective, which may be the training time, the training energy consumption, the model accuracy and so on. Then we can formulate the target multi-objective HPO problem as follow: 
\begin{equation}
\label{eq:1}
\begin{aligned}
      arg\min \boldsymbol{f}(\boldsymbol{x}), \boldsymbol{x} \in \mathbb{X}
\end{aligned}
\end{equation}
where $ \boldsymbol{f}(\boldsymbol{x})=\left\{f_{1}(\boldsymbol{x}), f_{2}(\boldsymbol{x}), \ldots, f_{m}(\boldsymbol{x}), \ldots, f_{M}(\boldsymbol{x})\right\} $. Noting that for a multi-objective optimization problem, it is generally not possible to find a solution that can simultaneously optimize each objective $f_m$ due to the conflicting nature of these objectives. Instead, a feasible solution is
exploring the candidate configuration space $\mathbb{X}$ to find a set of non-dominated configurations which equally efficient trade-offs among the objectives. Specifically, we say that configuration $\boldsymbol{x}$ Pareto-dominates another configuration $\boldsymbol{x'}$ if it satisfies: 
\begin{equation}
\label{eq:2}
\left\{\begin{array}{l}
f_{m}(\boldsymbol{x}) \leq f_{m}(\boldsymbol{x'}), \forall m \in\{1,2, \ldots, M\} \\
f_{j}(\boldsymbol{x})<f_{j}(\boldsymbol{x'}), \exists j \in\{1,2, \ldots, M\}
\end{array}\right.
\end{equation}
Then the optimal solution to a multi-objective optimization problem is actually a set of configurations $\boldsymbol{X^*} \subset \mathbb{X}$, which satisfies that no configuration $\boldsymbol{x'} \in\mathbb{X} \setminus \boldsymbol{X^*}$ can Pareto-dominate a configuration $ \boldsymbol{x} \in \boldsymbol{X^*}$. Without loss of generality, the solution set $\boldsymbol{X^*}$ is called the Pareto-optimal set and the corresponding set of function values is called the Pareto front. 

\begin{figure*}[t]
\label{fig}
\centerline{\includegraphics[width=17cm,height=7cm]{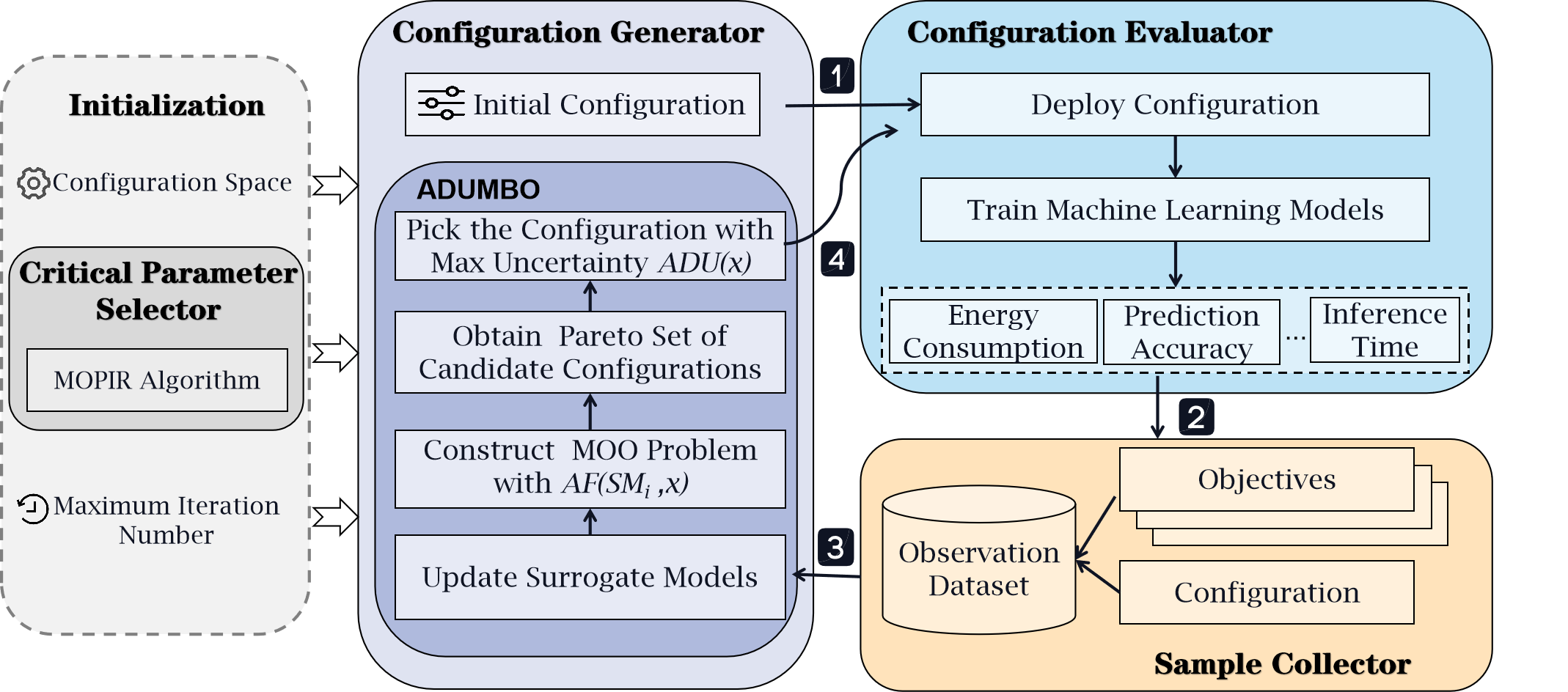}}
\caption{The architecture overview of HyperTuner.}
\label{fig:3}
\end{figure*}

\textbf{Challenges and Limitations of Prior Methods.} In fact, the objective functions are usually black-box and we can only obtain the corresponding function values through practical configuration evaluation. Considering the non-negligible evaluation cost in terms of time, energy or resources for each candidate configuration, the Pareto-optimal set $\boldsymbol{X^*}$ for Problem \ref{eq:1} should be achieved effectively as well as efficiently. Recently, due to its sample-efficiency feature for black-box optimization problems, the vanilla Bayesian optimization algorithm (BO) \cite{bosuperior} has been extended to deal with multiple objectives to achieve a good Pareto-optimal set approximation within a constrained number of configuration evaluations. However, as described above, the cross-layer system parameters and model hyperparameters related to different optimization objectives together form a complex high dimensional configuration space. As a result, none of prior BO-based methods is able to effectively and efficiently solve the formulated multi-objective black-box optimization problem. For example, PESMO \cite{hernandez2016predictive} is a state-of-the-art method based on entropy optimization, but it is computationally expensive to conduct the optimization process considering the high dimensional configuration space. Although FlexiBO \cite{iqbal2020flexibo} takes the computational cost into consideration, however, it assumes a discrete configuration space and traverses it for the next candidate configuration. Therefore, in this paper, we aim to properly solve the formulated multi-objective black-box optimization problem as described in Problem 1 in order to achieve cross-layer multi-objective hyperparameter optimization for data analytic services.

\section{System Overview}
\label{sec:3}
We propose a new black-box approach named HyperTuner in this paper, which is designed to achieve multi-objective hyperparameter optimization for data analytic services via auto-tuning their cross-layer configuration parameters. To achieve this target, we have to solve the formulated black-box multi-objective optimization problem described as Problem 1. The key idea of HyperTuner is to first conduct multi-objective parameter importance ranking with our MOPIR algorithm and then leverage the proposed ADUMBO algorithm to tackle with this challenging problem. We will thoroughly describe these two algorithms in next section. 
Fig \ref{fig:3} shows a high-level overview of HyperTuner. Specifically, HyperTuner mainly consists of four modules, which are respectively Critical Parameters Selector, Configuration Generator, Configuration Evaluator  and Sample Collector. We give a brief introduction of each module as following:

\textbf{Critical Parameter Selector} is responsible for pick out critical parameters through parameter importance ranking to address the challenge of high-dimensional configuration space. To this end, we design MOPIR, which is able to obtain the Pareto importance rank of each parameter based on their Gini index along each objective as well as the non-dominated sorting method. For a certain data analytic service, critical parameter selection is usually a one-time cost but can effectively shrink the high-dimensional configuration space consisting of cross-layer parameters from the whole machine learning system stack. Detailed description of MOPIR
can be found in Sec. \ref{sec:4a}.

\textbf{Configuration Generator} is responsible for iteratively generating a new configuration based on current observation dataset. The core lies in this module is the proposed ADUMBO algorithm. Inspired by the recent work USeMO \cite{belakaria2020uncertainty}, ADUMBO is able to efficiently solve the formulated black-box
multi-objective optimization problem via introducing a novel
adaptive uncertainty metric. Different from prior methods, ADUMBO only needs to solve a cheap multi-objective optimization problem defined by the acquisition functions and leverages the adaptive uncertainty metric to guide the choice of next promising configuration to be evaluated during each iteration. Detailed description of ADUMBO
can be found in Sec. \ref{sec4b}.

\textbf{Configuration Evaluator} is responsible for evaluating the model training metrics related to user-specified optimization objectives under the newly generated configuration from the Configuration Generator module. To this end, HyperTuner will first update the configuration of the whole underlying machine learning system stack and then execute the target model training tasks again to obtain its corresponding runtime metrics. These metrics are usually good indicators of the optimization objectives. It is worth noting that practical testbed may contain metric noises such as network jitters for distributed training. In order to mitigate
the impact of metric variability, Configuration Evaluator re-executes the model training tasks for a certain times for each configuration. 

\textbf{Sample Collector} is responsible for collecting the model training metrics generated by the Configuration Evaluator module and calculating the corresponding values of the $M$ optimization objectives under the newly generated configuration $x$. After that, it adds the new observation sample $<$$x$, $obj\_i$$>$ ($i=1,2,...M$) into the observation dataset. The surrogate model for each objective will be updated according to the newest observation dataset in the Configuration Generator module and launch the next optimization iteration. 

For ease of understanding, we briefly describe how to leverage HyperTuner to auto-tune the model hyperparameters as well as system parameters of data analytic services for multi-objective optimization. Once the configuration optimization process is necessary to be launched, developers should first pick out important parameters for their optimization objectives using the Critical Parameter Selector and then specify the candidate configuration space as well as the maximum iteration number. After that, Configuration Generator, Configuration Evaluator and Sample Collector will collaborate with each other in an iterative manner within the specified iteration number constraint. When the total number of observation samples reaches the constraint, Hypertuner will terminate and report the Pareto-optimal set of configurations ever found. 

\section{Design and Implementation of HyperTuner}
\label{sec:4}

In this section, we first introduce our parameter importance ranking method MOPIR tailored for multiple objectives. Then we describe the details of the proposed ADUMBO algorithm in HyperTuner, which is able to adaptively leverage the uncertainty across surrogate models of each objective to address the formulated multi-objective black-box optimization problem. Finally, we also give a short description of the practical implementation of HyperTuner. 

\subsection{MOPIR: Parameter Importance Ranking for Multiple Objectives}
\label{sec:4a}

Parameter importance ranking is the most common method to pick out critical parameters to tackle with the challenge of high-dimensional configuration space. The multiple optimization objectives in our problem further compound this challenge since the importance of parameters is quite different under different optimization objectives. There are several widely used methods such as Lasso \cite{lasso} and Gini index \cite{gini} can be applied for single-objective parameter importance ranking, however, it is still a challenging task to choose critical parameters for the multi-objective configuration auto-tuning problem based on the single-objective ranking results. For example, it is common that one parameter ranks in a quite different order under different objectives. In order to determine the rank of a certain parameter across multiple objectives, a straightforward way is to distribute a weight to each order under each objective and then calculate a weighted order for this parameter. Unfortunately, the weight distribution work is not easy since the importance of each objective may vary with different data analytic services. 

To address this challenge, we design MOPIR (short for Multi-Objective Parameter Importance Ranking) in this paper, which is able to obtain the Pareto rank of each parameter across all the user-specified optimization objectives. In detail, MOPIR first leverages the single-objective Gini index method to obtain the Gini index $GI^{m}_{i}$ of each candidate parameter $x_i$ ($i=1,2,...,D$) under objective $m$ ($m=1,2,...,M$). After that, MOPIR calculates the Pareto rank of each parameter $x_i$ based on the non-dominanted sorting method. It is worth noting that the importance of parameters with higher Pareto rank dominate the lower ones while parameters with same Pareto rank do not dominate each other. In order to pick $d$ ($d<<D$) critical parameter out from the original $D$ parameters to address the challenge of high-dimensional configuration space, MOPIR select parameters from the highest Pareto rank and then the lower rank. If we have to select part of the parameters from the same Pareto rank, MOPIR prefers the one with maximum Gini index along a certain optimization objective. 
\begin{figure}[h]
    \centering
    \includegraphics[width=7cm,height=6cm]{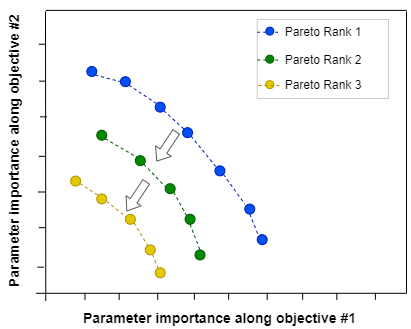}
    \caption{The Pareto rank of candidate parameters under a bi-objective optimization scenario. }
    \label{fig:4.1}
\end{figure}
\subsection{ADUMBO: Multi-Objective Bayesian Optimization with Adaptive Uncertainty}
\label{sec4b}

Considering the complex relationship between the objective function and the cross-layer configuration parameters, it is impossible to catch the accurate prediction model for each objective. That is to say, the objective functions are usually black-box and we can only obtain the corresponding function values through expensive configuration evaluation. Therefore, besides carefully selecting important parameters for the target multiple objectives, we also propose an ADaptive Uncertainty-based Multi-objective Bayesian Optimization algorithm named ADUMBO for HyperTuner. To sum up, ADUMBO is inspired by a recent work USeMO and further improves the performance in solving black-box multi-objective optimization problems via introducing a novel adaptive uncertainty metric. As the same with vanilla Bayesian optimization algorithm, ADUMBO also works in an iterative manner which involves three key steps: 1) Update the surrogate model for each objective function based on the latest dataset of configuration evaluations; 2) Solve a cheap multi-objective optimization problem defined by the corresponding acquisition functions to generate a Pareto-optimal set of candidate configurations. Different from the USeMO algorithm, ADUMBO then selects the most promising candidate configuration from the generated Pareto-optimal set via maximizing a new well-designed metric which can adaptively leverage the uncertainty as well as the predicted mean across all the surrogate models along with the iteration times; 3) Evaluate the newly generated configuration in HyperTuner and start the next iteration. 

The detail description of ADUMBO can be found in Algorithm \ref{alg1}. In the initialization stage, ADUMBO randomly selects $k$ configurations from the candidate configuration space and evaluate each configuration to obtain the initial observation dataset $\mathbb{D}$, and then builds surrogate models $SM_i$ for each objective $O_i$ ($i=1,2,...,M$) based on these initial samples (Line 1-2). We use the Gaussian Process in this paper and leave other surrogate models as the future work. After that, ADUMBO iteratively generates the most promising configurations if current evaluation times is less than the user-specified constraint $T_{max}$. Specifically, given an acquisition function $AF$ for each surrogate model, a cheap multi-objective optimization problem with objectives $AF(SM_1,\boldsymbol{x})$, $AF(SM_2,\boldsymbol{x})$, $\dots$, $AF(SM_M, \boldsymbol{x})$ is constructed and the non-dominated sorting genetic algorithm NSGA-II\cite{nsga} is utilized to solve this problem to obtain a Pareto-optimal set of configurations for evaluation $\boldsymbol{X_p}$ (Line 4). However, considering the non-ignorable configuration evaluation costs, it is still necessary to determine the most promising one from $\boldsymbol{X_p}$ (Line 5-6). To this end, we design a new uncertainty metric $ADU$ which can be well adaptive to current iteration times. In detail, the original uncertainty metric in USeMO algorithm is defined as the volume of the uncertainty hyper-rectangle across all the surrogate models: 
\begin{equation}
   U_{\beta_t}(\boldsymbol{x})=\operatorname{VOL}\left(\left\{L C B\left(SM_{i}, \boldsymbol{x}\right), U C B\left(SM_{i}, \boldsymbol{x}\right)\right\}_{i=1}^{M}\right)
\end{equation}
where $LCB(\boldsymbol{x})=\mu(\boldsymbol{x})-{\beta_{t}}^{1 / 2}\sigma(\boldsymbol{x})$, $UCB(\boldsymbol{x})=\mu(\boldsymbol{x})+{\beta_{t}}^{1 / 2}\sigma(\boldsymbol{x})$, and $\beta_t=2 \log (\left | \mathbb{X}  \right |  \pi^2 t^2 / 6 \delta)$ is the parameter which aims to tradeoff between the mean $\mu(\boldsymbol{x})$ (exploitation score) and the standard deviation $\sigma(\boldsymbol{x})$ (exploration score) of the the surrogate model for current iteration $t$. Unfortunately, according to the current implementation of USeMO, we can obtain that:
\begin{equation}
\label{eq:4}
 \begin{aligned}
U_{\beta_t}(\boldsymbol{x})&= \prod_{i=1}^{M} [U C B(SM_i, \boldsymbol{x})-L C B(SM_i, \boldsymbol{x})]\\
    &=\prod_{i=1}^{M} \beta_{t}^{1 / 2} \sigma(SM_i,\boldsymbol{x})\\
\end{aligned}
\end{equation}

which tells the truth that the metric used in USeMO only cares  about the variances of current surrogate models and the parameter $\beta_t$ fails to help achieve the exploitation-exploration balance. In fact, the surrogate model approaches closer to the real objective function with the iteration number increasing and thus it is not wise enough to choose the configuration for practical evaluation only regarding to above uncertainty metric. Instead, we insist that more attention should be paid to predicted mean of each surrogate model. Specifically, the adaptive uncertainty metric $ADU_{\beta_{t}}(\boldsymbol{x})$ for current iteration $t$ is denoted as:
\begin{equation}
 \begin{aligned}
    ADU_{\beta_{t}}(\boldsymbol{x}) = \beta_{t}^{1 / 2} \prod_{i=1}^{M} \mu(SM_i,\boldsymbol{x}) + \prod_{i=1}^{M}  \sigma(SM_i,\boldsymbol{x})\\ 
\end{aligned}
\end{equation}
where $\beta_t=2 \log (\left | \mathbb{X}  \right |  \pi^2 t^2 / 6 \delta)$ is able to balance the exploitation and exploration across all the surrogate models adaptively under iteration $t$. Next, the selected configuration $\boldsymbol{x}_{t+1}$ will be evaluated by HyperTuner to obtain the practical values of each objective function $f_{i}(\boldsymbol{x}_{t+1})$ and this new sample ($\boldsymbol{x}_{t+1}$, $\boldsymbol{f}(\boldsymbol{x}_{t+1})$) will be added to the observation dataset. Then the surrogate models for each objective function will be updated with the latest dataset and ADUMBO starts the next iteration (Line 7-10). When the given iteration times constraint is exhausted, ADUMBO will report the Pareto-optimal set of ever evaluated configurations for the target $M$ objectives. 
Considering the formulated multi-objective black-box optimization problem as described in Problem \ref{eq:1}, ADUMBO does not have to traverse the whole candidate configuration space to generate the next most promising configuration for evaluation. Besides the sample-efficiency feature, ADUMBO itself is also a computationally-efficient algorithm compared with prior methods since only a cheap MO optimization problem need to be solved during each iteration. Therefore, leveraging MOPIR and ADUMBO, HyperTuner is able to well address the associated three challenges to achieve cross-layer multi-objective hyperparameter tuning for data analytic services. 
\begin{algorithm}[htbp]
        \caption{ADUMBO Detail}
 \label{alg1}
 {\bf Input:} Candidate configuration space $\mathbb{X}$; Maximum iteration number $T_{max}$; $M$ Black-box objective functions $f_{1}(x)$, $f_{2}(x)$, \dots, $f_{M}(x)$.\\
 {\bf Output:} Pareto-optimal set $\boldsymbol{X^*}$ and Pareto front of evaluated configurations for $M$ objectives.\\
 \begin{algorithmic}[1] 
  \STATE Randomly select $k$ samples from $\mathbb{X}$ and evaluate each configuration to obtain the initial observation dataset $\mathbb{D}$; 
  \STATE Initialize surrogate models $SM_1$, $SM_2$, \dots, $SM_M$ for each objective from $\mathbb{D}$; 
  \WHILE {Current iteration $t \le T_{max}$}
    \STATE Use NSGA-II to solve the multi-objective optimization problem with acquisition functions $AF(SM_1, \boldsymbol{x})$, $AF(SM_2, \boldsymbol{x})$, $\ldots$, $AF(SM_M, \boldsymbol{x})$ to obtain a Pareto-optimal set of candidate configurations $\boldsymbol{X_p}$;
    \STATE Compute the adaptive uncertainty metric for each $\boldsymbol{x}\in \boldsymbol{X_p}$: $ADU(\boldsymbol{x}) =  \prod_{i=1}^{M}  \sigma(SM_i,\boldsymbol{x}) +\beta_{t}^{1 / 2} \prod_{i=1}^{M} \mu(SM_i,\boldsymbol{x})$;
     \STATE Pick the candidate configuration with maximum  uncertainty metric: $\boldsymbol{x_{t+1}} = arg max_{ \boldsymbol{x} \in \boldsymbol{X_p}} ADU(\boldsymbol{x})$;
     \STATE Evaluate the selected configuration $\boldsymbol{x_{t+1}}$ to obtain practical objective values: $\boldsymbol{f}(\boldsymbol{x_{t+1}})=(f_1(\boldsymbol{x_{t+1}}), f_2( \boldsymbol{x_{t+1}}), \dots, f_M(\boldsymbol{x_{t+1}}))$;
     \STATE Add the new sample to the observation dataset $\mathbb{D}_{t+1}=\mathbb{D}_{t} \bigcup (\boldsymbol{x_{t+1}},\boldsymbol{f}(\boldsymbol{x_{t+1}}))$;
    \STATE Update surrogate models $SM_1$, $SM_2$, \dots, $SM_M$  with the latest dataset $\mathbb{D}_{t+1}$; 
    \STATE $t = t+1 $;
\ENDWHILE
 \end{algorithmic}
 \end{algorithm}
\subsection{Implementation of HyperTuner}
\label{sec:4c}
To achieve cross-layer multi-objective hyperparameter auto-tuning for data analytic services, besides iteratively generating new promising configuration with ADUMBO algorithm, HyperTuner should also be able to automatically deploy configuration updates and launch model training tasks on the local distributed machine learning cluster. Towards this target, we use Python to implement the major part of HyperTuner and 
leverage the interfaces provided by the Ansible tool \cite{ansibleplaybook} to achieve automatic execution. On the one hand, configuration parameters from different layer usually have quite different update methods. For instance, while kernel parameters can be directly updated through \emph{sysctl -p} command, while a machine learning framework based on Spark has to modify the corresponding configuration file and then restart the whole cluster to put the configuration change into effect. Therefore, we compose dedicated Ansible playbooks in YAML syntax to automate this complex configuration update procedure in HyperTuner. On the other hand, different model training tasks from the data analytic services should be automatically executed to evaluate the corresponding metrics of user-specified optimization objectives under the newly generated configuration. Again, we compose dedicated Ansible playbooks to automatically launch model training tasks and collect the associated system metrics.With these  Python codes and Ansible playbooks, HyperTuner is able to jointly auto-tune the cross-layer model hyperparameters as well as system parameters and report the Pareto-optimal configuration set found within the specified iteration times constraint. 

It is worth noting that HyperTuner can be easily extended to support different machine learning platforms and different optimization objectives. To evaluate the adaptability of HyperTuner, we conduct experiments and make a discussion based on experimental results in Sec. \ref{sec:6d}. Finally, we open source the implementation of HyperTuner at https://github.com/zss233-21/HyperTuner.git for more potential practical applications.

\section{Experimental Setups}
\label{sec:5}
\subsection{Experimental Platform}

To evaluate the effectiveness and efficiency of HyperTuner for data analytic services, we conduct a series of experiments on a local cluster consisting of three Linux servers. All the servers are connected with LAN and equipped with an Intel(R) Core(TM) i7-10700 with 8 physical cores, 16 GB DDR4 memory and 1 TB HDD. The chosen operating system is Ubuntu 20.04.3 LTS with the 5.11.0-41-generic kernel version. Besides, TensorFlow 2.7 with CPU backend is deployed on each server and provides distributed training via API \emph{tf.distribute.Strategy}. Python and Keras API are leveraged to implement the deep neural network architectures in our experiments. Specifically, we also make sure that nothing else is running except the essential kernel processes and the target model training tasks in order to mitigate the potential interference. We also conduct experiments on a local distributed Spark cluster with several machine learning applications to evaluate the adaptability of HyperTuner and detail discussions can be found in Sec.\ref{sec:6d}.

\subsection{Data Analytic Jobs and Tuning Objectives}
As described in Table \ref{tab1}, we consider three data analytic jobs composed of training representative deep learning models LeNet, VGG-9 and LSTM respectively on three different datasets \emph{MNIST}\cite{lenet}, \emph{Cats vs. Dogs}\cite{catvsdog} and \emph{IMDB\_Sentiment}\cite{imdb}. Specifically, the selected models can be applied to different scenarios including image recognition and machine translation. In addition, we cut the \emph{Cats vs. Dogs} dataset in half considering the limitation of available computing resources as well as the total experimental time consumption. For each training task, we regard the model training time and the associating energy consumption as the tuning objectives. Besides, the model accuracy must surpass a pre-specified threshold. It is worth noting that since the power fluctuation range of other components in a server is relatively a small value, we only measure the energy consumption of the whole CPU package for each second utilizing the turbostat toolkit\cite{turbostat} and then calculate the total energy consumption during the model training process. Finally, to evaluate the adaptability of HyperTuner, we also conduct experiments under other model training tasks as well as different tuning objectives and detail discussions can be found in Sec. \ref{sec:6d}.
\begin{table}[htbp]
\caption{The DNN models and their datasets.}
\scalebox{0.8}{
    \begin{tabular}{|c|c|c|c|c|}
    \hline
    \textbf{Model}&\textbf{Datasets}&\textbf{Training Size}&\textbf{Test Size}&\textbf{Domain} \\
    \hline
    LeNet&MNIST	&60,000	&10,000	&Image \\
    \hline
    VGG-9&Cats vs. Dogs&25,000&12,500&Image \\
    \hline
    LSTM&IMDB Sentiment&25,00d0&25,000&NLP \\
    \hline
    \end{tabular}
    }
\label{tab1}
\end{table}

\subsection{Cross-Layer Configuration Parameters}
For the user-specified tuning objectives, we utilize the MOPIR method to rank the importance of parameters and then select 15 critical parameters in total across the whole system stack. As described in Table \ref{tab2}, there are 3 model hyperparameters and 12 system parameters including those from the Tensorflow framework, kernel and the CPU backends. We also list the value range as well as the detail description of each chosen parameter. On the one hand, it is worth noting that due to the lack of official guidance information, the hyperparameter value ranges are settled according to empirical information from our practical experiments as well as other public experiences. On the other hand, the selected kernel parameters are closely related to the performance of interprocess communication, virtual memory usage and disk I/O. According to the official guides and the hardware limitations, we set the value range of each kernel parameter. In addition, the value ranges of Tensorflow parameters \emph{inter\_op\_parallelism\_threads} and \emph{intra\_op\_ parallelism\_threads} are determined according to \cite{tf}. 

\begin{table*}[h]
\centering
\caption{The cross-layer critical parameters for DNN model training tasks.}
\begin{center}
\scalebox{0.8}{
\begin{tabular}{|c|c|c|l|}
\hline
\textbf{Type}&\textbf{Parameter} &\textbf{Range} &\multicolumn{1}{c|}{\textbf{Description}}\\
\hline
\textbf{}&\textit{learning\_rate} & [0.01,0.001, 0.0001, 0.00001]&The step size for gradient descent \\
\cline{2-4}  
\textbf{HyperParameter}& \textit{batchsize} &[32, 64, 96, 128]&The number of samples per gradient update \\
\cline{2-4}  
\textbf{}&\textit{dropout\_rate}&	[0.4, 0.5, 0.6, 0.7, 0.8]	&The fraction of a layer’s neurons be zeroed out\\
\cline{1-4}  
\multirow{2}{*}{\textbf{Tensorflow}}&\textit{inter\_op\_parallelism\_threads}&[1, 2, 3, 4]&\makecell[l]{Maximum number of independent computations\\to execute in parallel}\\
\cline{2-4}
&\textit{intra\_op\_parallelism\_threads}&	[14, 21, 28, 35, 42, 49, 56]&\makecell[l]{Maximum number of threads to use for executing a \\single computation} \\
\cline{1-4}
\multirow{9}{*}{\textbf{Kernel}}&\textit{swappiness}&	0-100&Frequency to drop caches and swap out application pages\\
\cline{2-4}
\textbf{}&\textit{dirty\_ratio}&	10-100&	 Percentage of dirty pages in memory\\
\cline{2-4}
\textbf{}&\textit{dirty\_background\_ratio}&	0-100&\makecell[l]{Max percentage of dirty pages before background writeback \\processes}\\
\cline{2-4}
\textbf{}&\textit{dirty\_expire\_centisecs}&	100-10000& Max duration interval of dirty pages\\
\cline{2-4}
\textbf{}&\textit{vfs\_cache\_pressure}&	50-150	&Frequency to retain dentry and inode caches\\
\cline{2-4}
&\textit{max\_map\_count}	&32765-98295	&Max number of mmap counts in memory\\
\cline{2-4}
\textbf{}&\textit{nr\_requests}&	64-256	&Number of requests allocated in block layer\\
\cline{2-4}
\textbf{}&\textit{read\_ahead\_kb}&	64-512	& How much extra data kernel reads from disk\\
\cline{2-4}
\textbf{}&\textit{somaxconn	}&128-4096& Max number of server's listening socket\\
\cline{1-4}
\textbf{CPU }& \textit{cpu\_ferq}	&\makecell{[800MHZ, 900MHz, 1.1GHz, \\ 1.2GHz, 1.4GHz, 1.5 GHz, 1.7GHz,\\1.8GHz, 2GHz, 2.2GHz, 2.4GHz,\\ 2.6GHz, 2.7 GHz, 2.9 GHz]} &\makecell[l]{Available frequency provided by the \textit{userspace} governor} \\
\cline{1-4} 
\end{tabular}
}
\label{tab2}
\end{center}
\end{table*}

\subsection{Baseline Algorithms}
\label{sec:5d}
The core of HyperTuner is the ADUMBO algorithm, which is proposed to efficiently and effectively solve the formulated black-box multi-objective optimization problem. To evaluate the performance of ADUMBO, we compare it with one single-objective optimization algorithm (Bayesian optimization\cite{bo}) and four multi-objective optimization HPO algorithms (Random search\cite{bergstra2012random}, PABO\cite{parsa2019pabo}, FlexiBO\cite{iqbal2020flexibo} and USeMO\cite{belakaria2020uncertainty}). Below we give a brief introduction of these baselines as well as how we implement them in HyperTuner: 
\begin{itemize}
\item \textbf{Bayesian optimization (BO)} is a sample-efficient black-box optimization algorithm. It utilizes a surrogate model to capture the prior distribution of the objective function and an acquisition function to generate the next observation point through trade-off between exploration and exploitation. Vanilla BO only supports a single objective and we use it to verify the necessity of multi-objective hyperparameter optimization. It is worth noting that we use Gaussian Process (GP) as the surrogate model and Expected Improvement (EI) as the acquisition function in our experiments. Besides, PABO, FlexiBO, USeMO and ADUMBO are all multi-objective extensions of vanilla BO and they share the same choice of surrogate model and acquisition function.
\item \textbf{Random search (RS)} generates configurations for evaluation from the candidate configuration space in a totally random manner for each iteration. With the same observation times constraint, it is more efficient than Grid search. However, since it does not take the relationship between parameters and the optimization objectives into account, it is not guaranteed to find a satisfying configuration even with a number of iterations. For each iteration, in order to generate a new random configuration, we use the \emph{randint()} function in Python to generate a random value for numerical parameters and a random index for categorical parameters.  
\item \textbf{PABO} is a multi-objective Bayesian optimization algorithm proposed recently for neural network hyperparameter optimization. The core insight of PABO is the pseudo-agent module and it works as follows: fits different objectives separately using Gaussian model and respectively generates different infill points for each objective; uses the pseudo-agent module to evaluate the effect of each recommended point on other optimization objectives; adds current point into the observation dataset of the corresponding surrogate model if it is not dominated by any historical points. Through this workflow, PABO can avoid solving the complex multi-objective optimization problem composed of acquisition functions and thus significantly reduce its computing time consumption.
\item \textbf{FlexiBO} is motivated by a multi-objective Bayesian optimization algorithm called Pareto Active Learning (PAL). It uses the same Pareto front construction method with PAL and introduces a cost-aware acquisition function to sample the next configuration for evaluation. During each iteration, FlexiBO constructs the optimistic Pareto front and pessimistic Pareto front using the uncertainty region for each candidate point in the configuration space. Then it selects the next configuration to evaluate according to the potential reduction in the uncertainty region between the pessimistic and optimistic Pareto fronts. Noting that a pre-defined evaluation cost under different objectives of the recommended configuration is also leveraged to determine which objective should be evaluated in practice. In our experiments, the candidate configuration space is not discrete and thus FlexiBO is not able to traverse the whole space to generate its cost-aware next configuration for evaluation. For fairness, we randomly select $|\boldsymbol{X_p}|$ points from the candidate configuration space, where $\boldsymbol{X_p}$ is the Pareto-optimal set of candidate configurations obtained by solving the multi-objective optimization problem with acquisition functions in our algorithm (see Sec.\ref{sec4b} for detail).  
\item \textbf{USeMO} is a new uncertainty-aware search framework that can efficiently generate configuration sequences to solve multi-objective  black-box optimization problems. The goal of USeMO is to approximate the true Pareto-optimal set while minimizing the necessary number of iterations. Specifically, it directly solves a cheap multi-objective optimization problem defined by the corresponding acquisition functions to generate a Pareto-optimal set of candidate configurations, and then identifies the best one according to a novel metric of uncertainty. However, this metric only cares about the variances of current surrogate models and fails to help achieve the exploitation-exploration balance. Hence, we propose an adaptive uncertainty metric in ADUMBO to better leverage the predicted mean of each surrogate model along with iterations. 
\end{itemize}

\section{Results and Analysis}
\label{sec:6}
\subsection{RQ1: Why multi-objective hyperparameter optimization is necessary for data analytic services?}
\begin{figure*}
 \centering
 \subfigure[]{    
  \label{fig:4a}\includegraphics[width=0.32\textwidth]{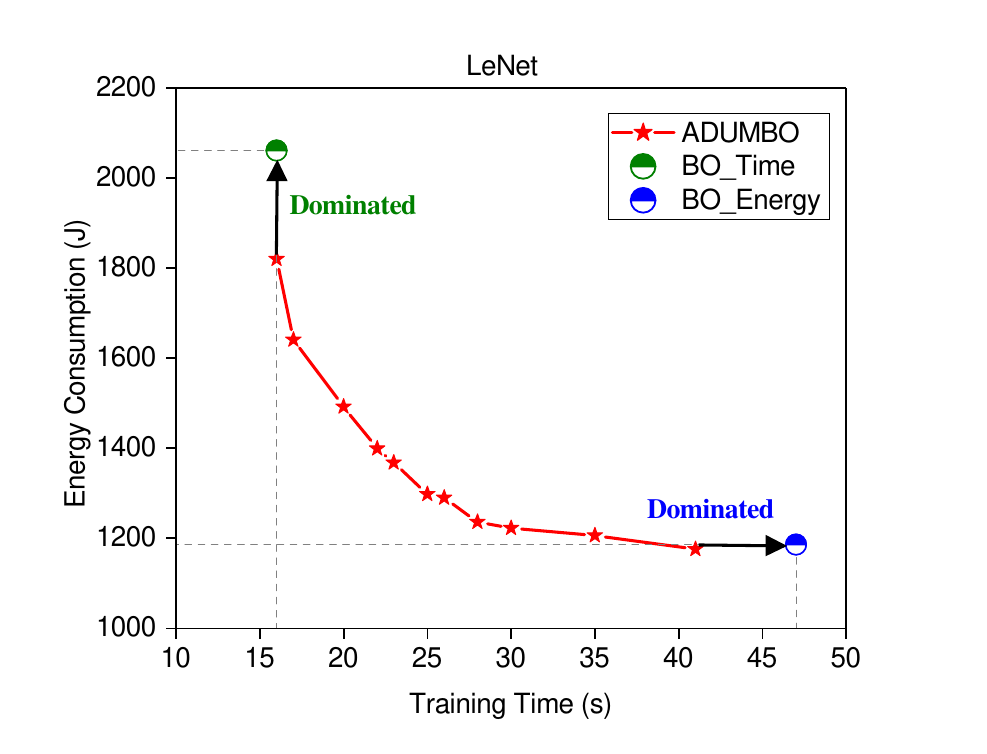}}
 \subfigure[]{  
  \label{fig:4b}\includegraphics[width=0.32\textwidth]{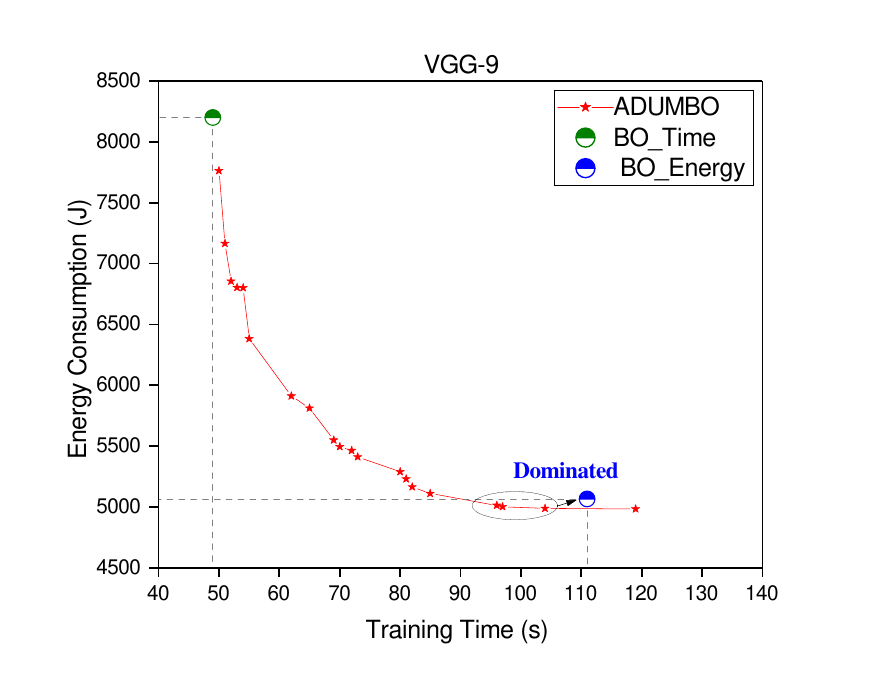}}
 \subfigure[]{
  \label{fig:4c}\includegraphics[width=0.32\textwidth]{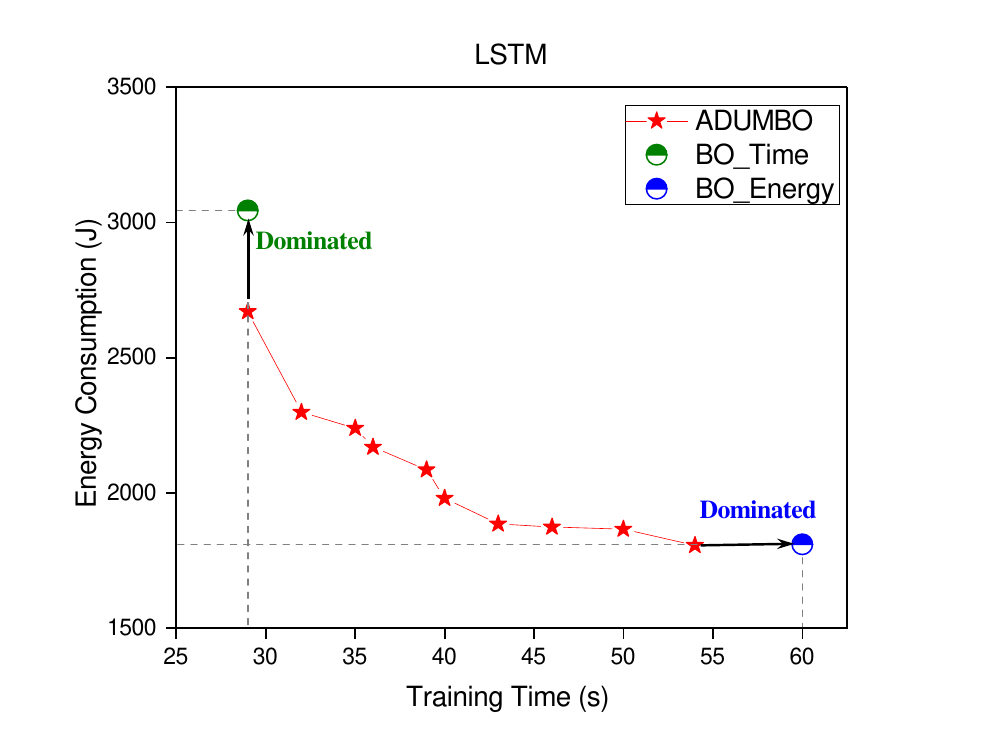}}
 \caption{The Pareto configuration set found by ADUMBO and optimal configurations found by two single-objective optimization algorithms.}
 \label{fig:4}
\end{figure*}
In this experiment, we aim to evaluate the limitations of conventional single-objective hyperparameter optimization algorithms when applied into auto-tuning the cross-layer configuration parameters for data analytic services with multiple optimization objectives. To this end, we implement ADUMBO and vanilla BO in HyperTuner and regard the model training time (in seconds) and energy consumption (in Joules) as the two optimization objectives. Specifically, BO\_Energy targets on energy reduction while BO\_Time targets on training time reduction. The total iteration times for ADUMBO, BO\_Time and BO\_Energy are all specified to 70. 

Figure \ref{fig:4} illustrates the Pareto configuration set found by ADUMBO as well as the two optimal configurations found by the two single-objective optimization algorithms for training three DNN models LeNet, VGG-9 and LSTM, respectively. We can find that ADUMBO is able to agilely tradeoff between model training time and energy consumption and almost always dominates the optimal configurations obtained by BO\_Energy and BO\_Time. For instance, although BO\_Energy is able to find a configuration with comparable total energy consumption, however, it consumes 14.63\%, 6.73\% and 11.11\% more model training time than ADUMBO respectively for the three DNN models. On the other hand, ADUMBO is able to find a configuration with close model training time while saves 11.65\%, 5.61\% and 12.83\% energy than BO\_Time. Noting that although the achieved model training time by ADUMBO is slightly larger (less than 2\%) than BO\_Time under VGG-9, the Pareto configuration set found by ADUMBO is still attractive considering the non-trivial reduction in the associated energy consumption. Therefore, it is essential to simultaneously take multiple optimization objectives into consideration and the proposed ADUMBO algorithm is always able to achieve a high-quality Pareto configuration set for different model training tasks. In addition, considering the difficulty to accurately measure the importance of different objectives\cite{overview}, transforming the original multi-objective HPO problem into a weighted single-objective one is not a promising solution either.

\subsection{RQ2:How does ADUMBO perform compared  with other multi-objective HPO algorithms?}
\label{sec:6b}
\begin{figure*}
 \centering
 \subfigure[]{    
  \label{fig:5a}\includegraphics[width=0.32\textwidth]{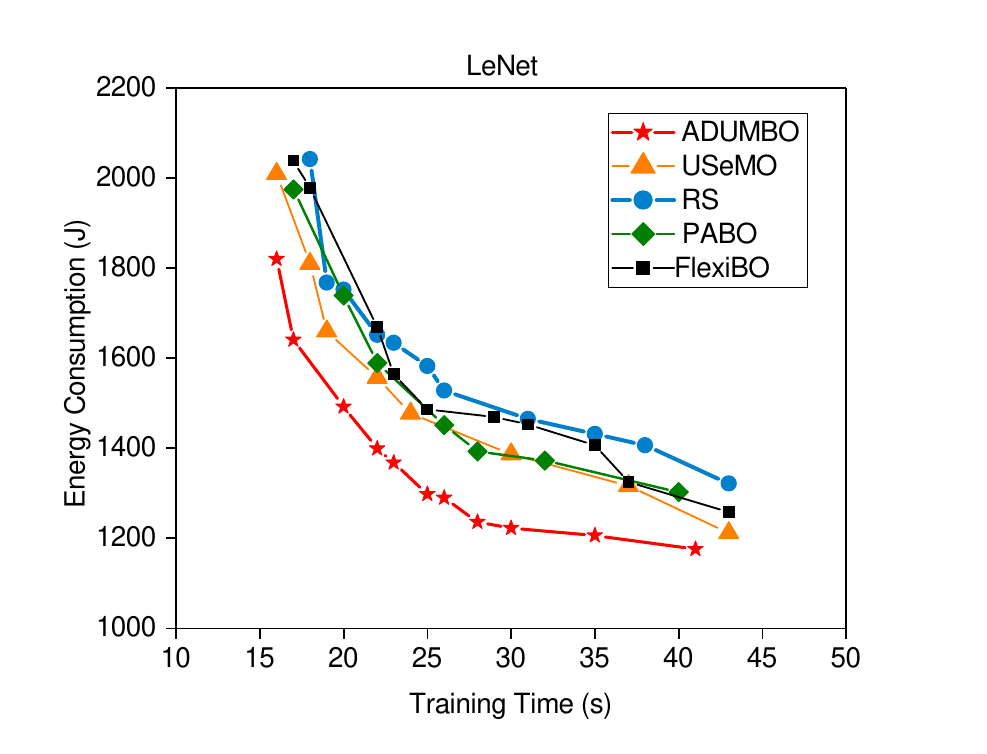}}
 \subfigure[]{  
  \label{fig:5b}\includegraphics[width=0.32\textwidth]{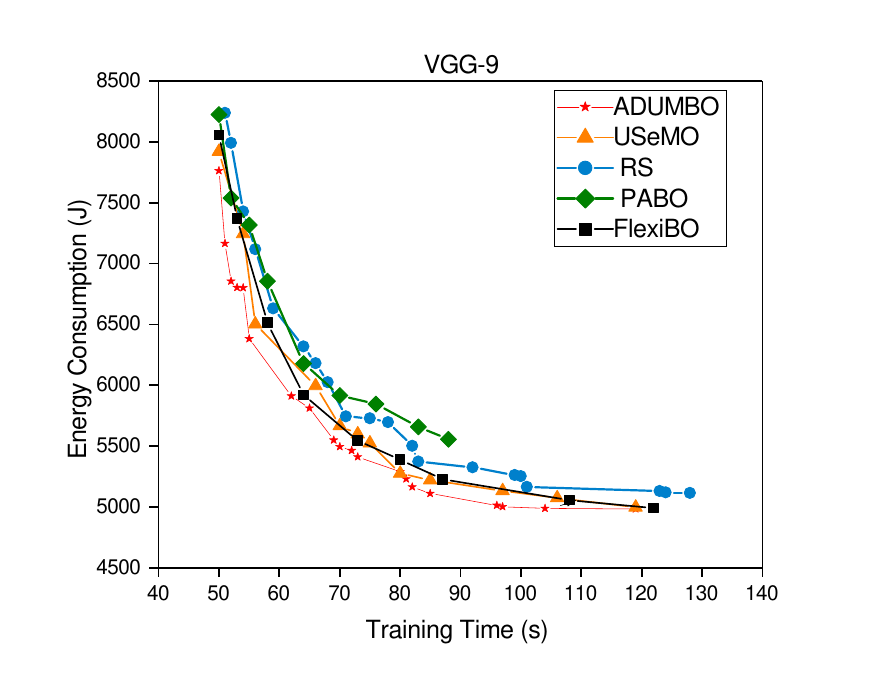}}
 \subfigure[]{
  \label{fig:5c}\includegraphics[width=0.32\textwidth]{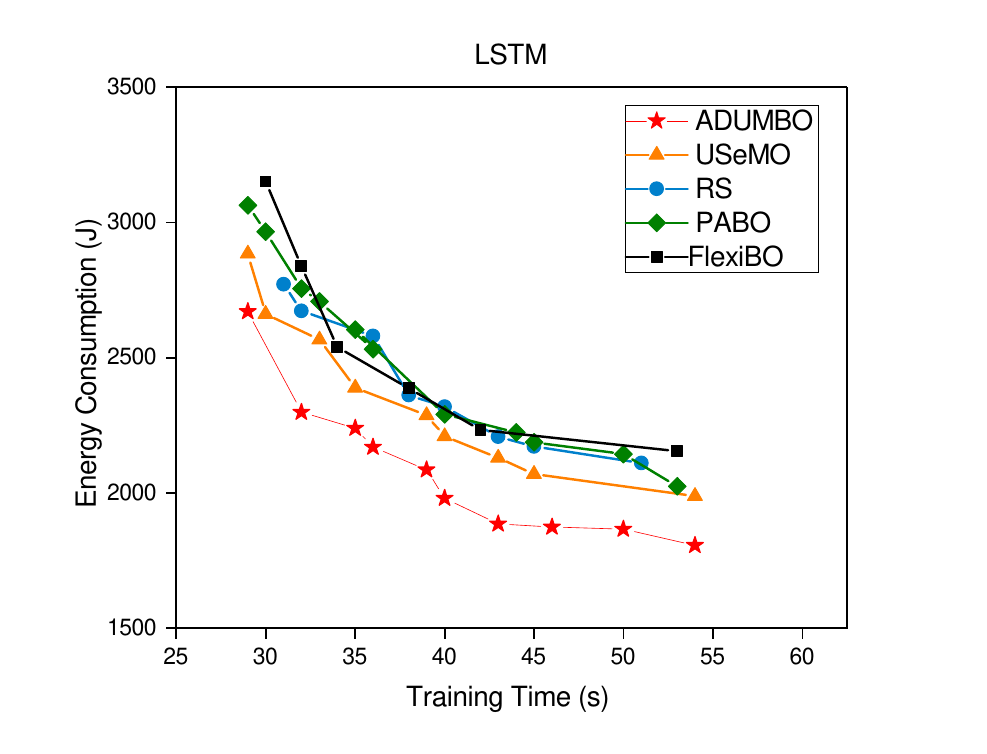}}    
 \subfigure[]{ \label{fig:5d}\includegraphics[width=0.32\textwidth]{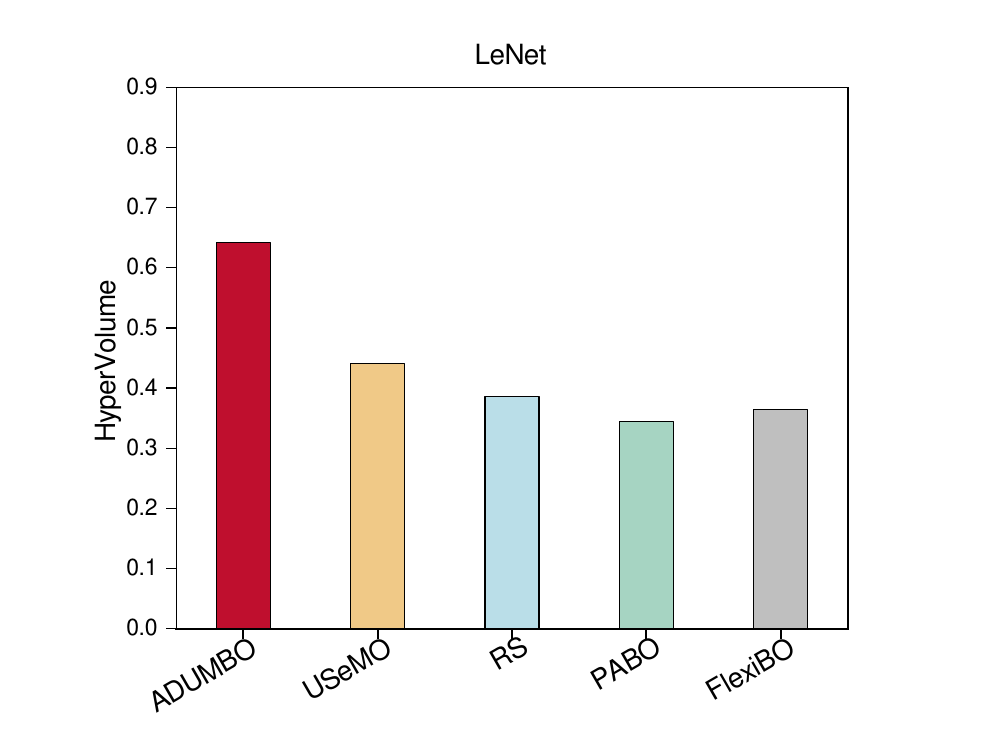}}
   \subfigure[]{   \label{fig:5e}\includegraphics[width=0.32\textwidth]{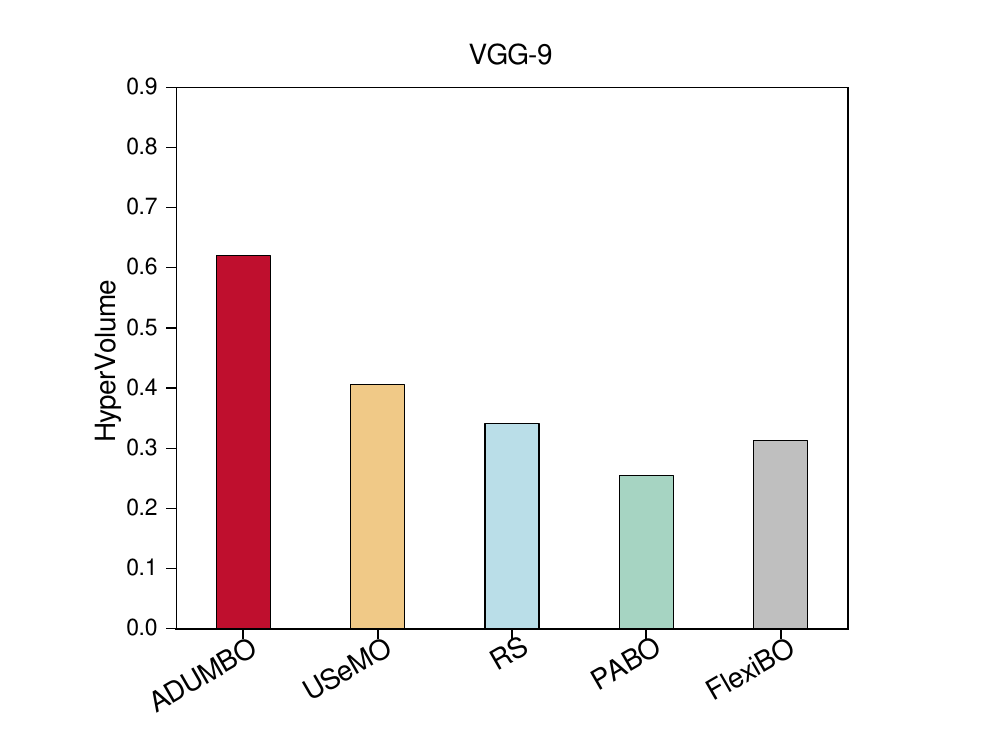}}  
   \subfigure[]{    
\label{fig:5f}\includegraphics[width=0.32\textwidth]{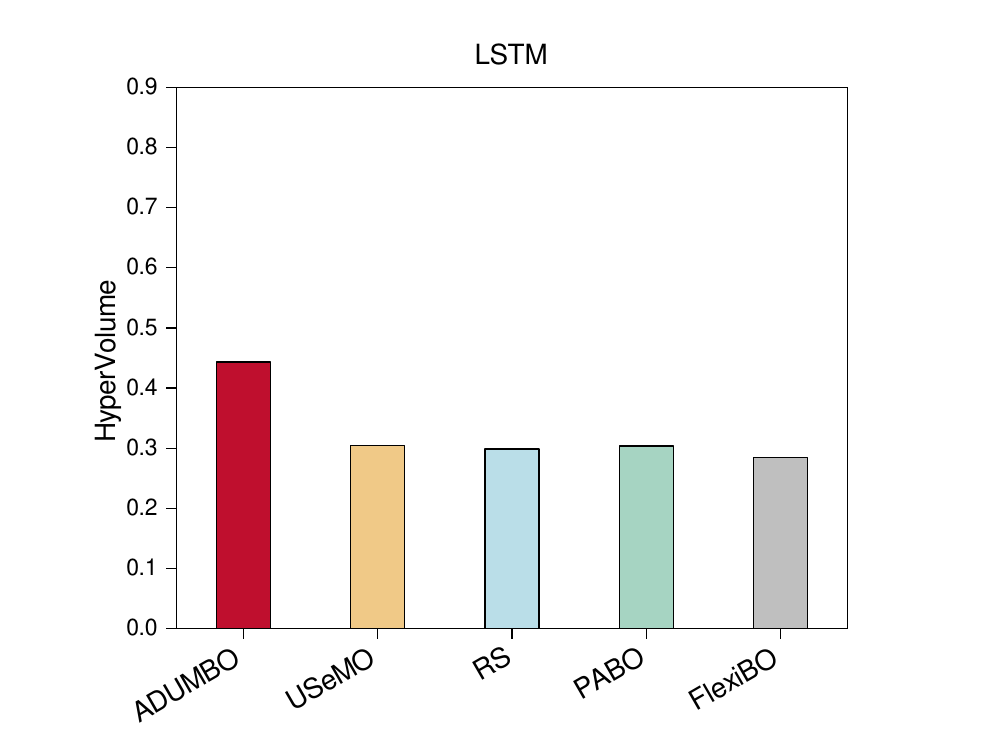}}
 \caption{Comparison of Pareto-optimal configuration set found by ADUMBO and four baseline algorithms.}
 \label{fig:5}
\end{figure*}
Above experimental results verify the necessity to conduct multi-objective hyperparameter optimization for data analytic services. In this section, to evaluate the performance of the proposed ADUMBO algorithm, we compare it with four baseline multi-objective HPO algorithms implemented in HyperTuner to tune the cross-layer parameters when training LeNet, VGG-9 and LSTM. The baseline algorithms are RS, PABO, FlexiBO and USeMO, respectively, and detail description can be found in Sec. \ref{sec:5d}. For all algorithms, the optimization objectives are the model training time and the associated total energy consumption. Besides, the number of total configuration evaluation times are all set to be 70. 

Figure \ref{fig:5a}-\ref{fig:5c} show the Pareto-optimal configuration set found by ADUMBO as well as the four baseline algorithms. It is obvious that ADUMBO is always able to find a better Pareto front for the three different DNN model training tasks. Significantly, the best training time achieved by ADUMBO under LeNet is 16 seconds and the corresponding energy consumption is 1820 Joules. Compared with RS, PABO, FlexiBO and USeMO, ADUMBO is able to save the total energy consumption respectively by 9.40\%, 7.81\%, 10.67\% and 9.41\% while consuming the least training time. To further give a deep quantitative analysis, we also leverage the HyperVolume ($HV$) indicator\cite{hypervolume} to measure the quality of the achieved Pareto fronts by different algorithms. Specifically, this indicator can simultaneously evaluate the convergence (minimizing the gap between candidate solutions and the true Pareto-optimal set) and diversity (maximizing the distribution of candidate solutions) of the achieved Pareto-optimal configuration set. Compared with other centralized indicators, the calculation of $HV$ indicator does not require a reference point from the real Pareto front, which is very difficult if not impossible to obtain in data analytic scenarios. In our experiment, we normalize the two objective values within [0, 1] and set (1.2, 1.2) as the reference point. The results are shown in Fig. \ref{fig:5d}-\ref{fig:5f}. In summary, the Pareto-optimal set of ADUMBO has higher convergence and diversity than all the other baselines. For instance, the $HV$ indicator of the Pareto front under LeNet achieved by ADUMBO is respetively 1.36$\times$, 1.69$\times$, 1.83$\times$ and 1.82$\times$ than that of USeMO, RS, PABO and FlexiBO. Similarly, the relative indicator achieved by ADUMBO ranges from 1.36$\times$ to 1.82$\times$ under LSTM. It is worth noting that the model structure of VGG-9 is relatively complex under our experimental hardware conditions and as a result, the optimization space is limited for the model training task. Even so, the $HV$ indicator of ADUMBO can still be as much as 2.01$\times$ compared with PABO. Therefore, the proposed ADUMBO algorithm is able to return a Pareto front with better convergence and diversity compared with prior multi-objective hyperparameter optimization algorithms. 

We can also find that USeMO achieves the most close performance compared with ADUMBO among the four baseline algorithms. In detail, USeMO first solves a cheap multi-objective optimization problem defined by the corresponding acquisition functions to generate a Pareto-optimal set of candidate configurations, and then identifies the best one according
to a novel metric of uncertainty. However, this metric only
cares about the variances of current surrogate models and totally neglects the predicted mean value of each surrogate model, which is actually more and more important along with iterations. Inspired by USeMO, we propose an adaptive uncertainty metric in ADUMBO to help 
fully leverage the predicted mean value and achieve better exploitation-exploration balance. Experimental results have verified the advantages of ADUMBO in solving multi-objective HPO problems for data analytic services.

\subsection{RQ3:Why jointly tuning cross-layer parameters is necessary for data analytic services?}
\begin{table*}[htbp]
\centering
\caption{Comparison of optimal configurations
found by Grid\_HP and ADUMBO under different optimization requirements.}
\begin{center}
\scalebox{0.9}{
\begin{tabular}{|c|c|c|c|c|c|c|c|c|}
\hline
\textbf{Model}& \textbf{Technique}& \multicolumn{2}{|c|}{\textbf{High Performance}} & \multicolumn{2}{|c|}{\textbf{Energy Conservation}}& \multicolumn{3}{|c|}{\textbf{Performance-Energy Balance}} \\
\cline{3-9}
\textbf{} & \textbf{}& \textbf{\textit{Runtime(s)}}& \textbf{\textit{Energy(J)}}& \textbf{\textit{Runtime(s)}}& \textbf{\textit{Energy(J)}}& \textbf{\textit{Runtime(s)}}& \textbf{\textit{Energy(J)}}&\textbf{\textit{Weighted Sum}} \\
\hline
LeNet & ADUMBO&\textbf{16}&	\textbf{1820}	&41&\textbf{1175}& 0.59 (22)		&0.44 (1398) &\textbf{0.52}\\
\cline{2-9} 
\textbf{} & Grid\_HP&19&	2232&	19 &	2232&	0.65(19) &	0.11(2232)& 0.38\\
\hline
VGG-9 & ADUMBO&\textbf{49}&	\textbf{7720}&	119	&\textbf{4997}&	0.60(55)& 0.66(6381)&\textbf{0.63}\\
\cline{2-9} 
\textbf{} & Grid\_HP&50&	16114&	73&9148& 0.51(66)&0.34(12432) &0.43	\\
\hline
LSTM & ADUMBO&\textbf{29}	&\textbf{2670}&	54 &	\textbf{1797}&	0.61(32)&0.51(2298)& \textbf{0.56}	\\
\cline{2-9} 
\textbf{} & Grid\_HP&30&	4420&	54	&3403&	0.55(37)&	0.26(3454)&0.41\\
\hline
\end{tabular}}
\label{tab3}
\end{center}
\end{table*}
As described in Sec. \ref{sec:2}, both model hyperparameters and system parameters
should be carefully tuned in order to conduct multi-objective hyperparameter optimization for data analytic services. To evaluate the necessity of jointly tuning these cross-layer parameters, in this experiment, we set the system parameters to their defaults and then utilize the Grid search method to tune the model hyperparameters when training LeNet, VGG-9 and LSTM. We denote above method as Grid\_HP and the detail description of these hyperparameters can be found in Table \ref{tab2}. As the same with ADUMBO (jointly tuning cross-layer parameters), we regard the model training time and the associated energy consumption as the two optimization objectives. 

Table \ref{tab3} compares the quality of the optimal configurations found by Grid\_HP and ADUMBO under different optimization requirements including high performance, energy conservation and performance-energy balance. Specifically, the optimal configuration of high performance or energy conservation requirements is the one with the shortest training time or the lowest energy consumption, respectively. For the balance requirement, the two objective
values were normalized to [0, 1] and the weight of each objective is set to be 0.5 and the configuration with lowest weighted sum is regarded as the optimal. We can find that through jointly tuning the cross-layer system parameters and model hyperparameters, ADUMBO is able to always find a better configuration under all the three different optimization requirements for all the three DNN model training tasks. Remarkably, under the high performance requirement, ADUMBO always spends the least model training time while saves the associated energy consumption by as much as 18.46\%, 52.09\% and 39.59\% respectively for the three DNN models. This significant improvement relies on the fact that the two optimization objectives are highly related to not only the model hyperparameters but also the system parameters, and these experimental results evaluate the necessity of jointly tuning the cross-layer parameters from the whole machine learning system stack. Besides, in order to address the accompanying challenge of high-dimensional candidate configuration space, HyperTuner leverages the MOPIR algorithm to select critical parameters for the target multiple objectives and then utilizes ADUMBO to explore a Pareto-optimal configuration set.

\subsection{RQ4:Whether the HyperTuner framework has enough adaptability to other optimization scenarios?}
\label{sec:6d}
\subsubsection{\textbf{Different datasets}}
\begin{figure*}
 \centering
 \subfigure[]{    
  \label{fig:7a}\includegraphics[width=0.32\textwidth]{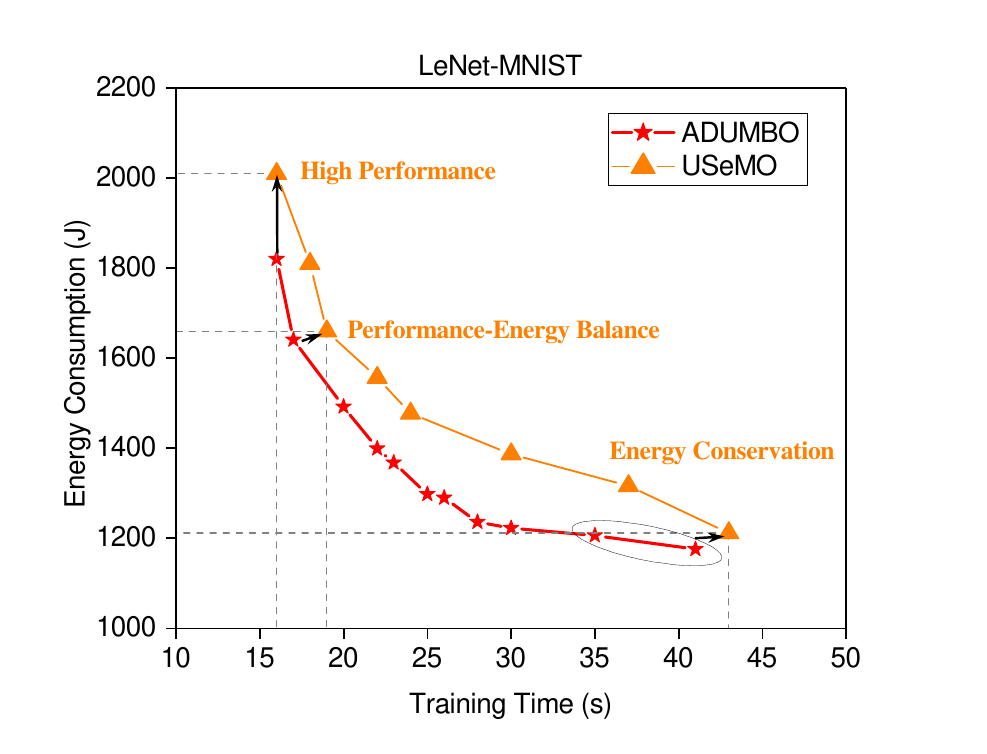}}
 \subfigure[]{  
  \label{fig:7b}\includegraphics[width=0.32\textwidth]{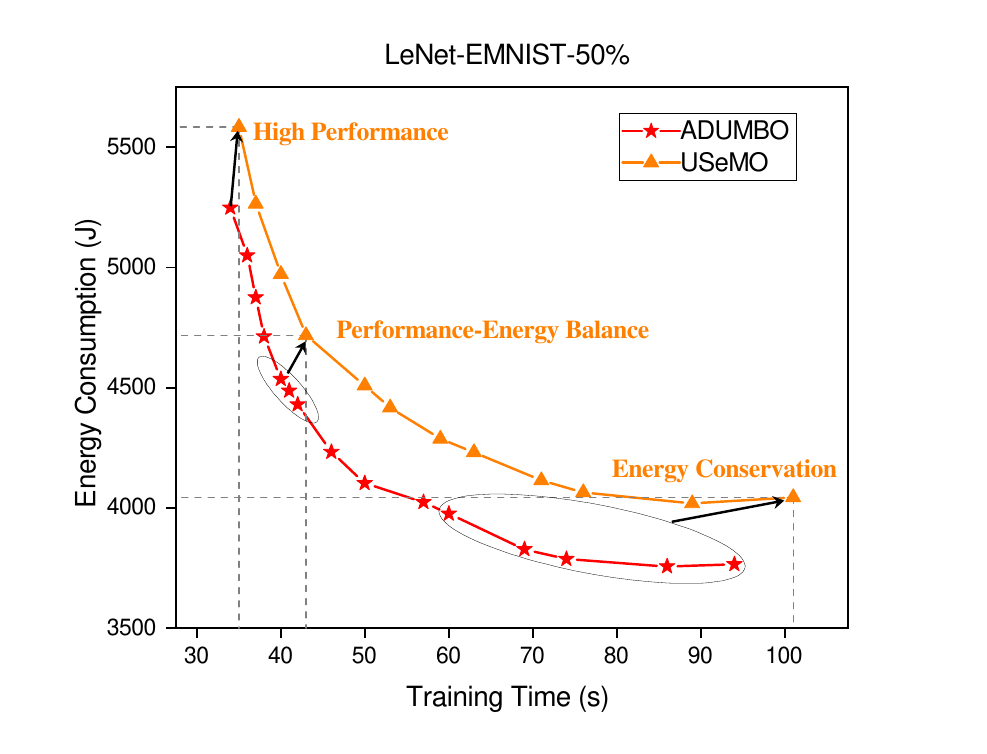}}
 \subfigure[]{
  \label{fig:7c}\includegraphics[width=0.32\textwidth]{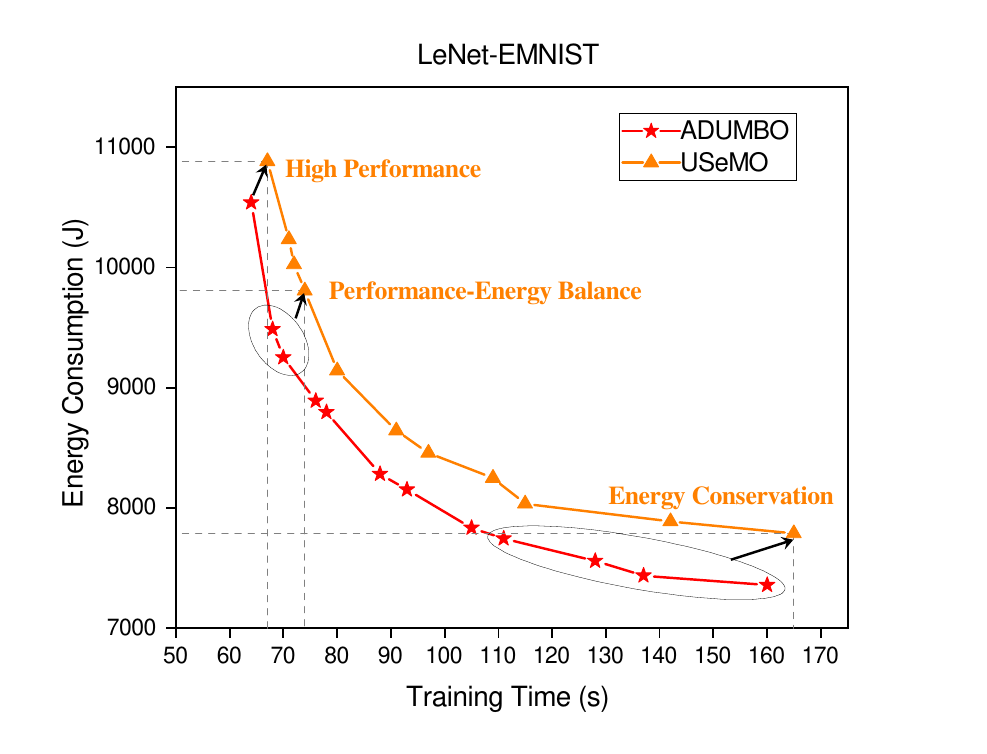}}
 \caption{The adaptability of HyperTuner to different model training datasets.}
 \label{fig:7}
\end{figure*}

We already conducted experiments to evaluate the performance of the proposed HyperTuner framework on training three classical DNN models with three different datasets listed in Table \ref{tab1}. In this experiment, we further evaluate the adaptability of HyperTuner under other dataset types and sizes. To this end, we introduced another dataset EMNIST (Expressed MNIST)\cite{emnist} whose training set includes 731,668 pictures and test set includes 82,587 pictures for handwritten font classification task. Specifically, we utilize HyperTuner to optimize the cross-layer parameters for LeNet model training tasks with MNIST, half of the EMNIST dataset and the full EMNIST dataset. The two optimization objectives are the model training time and the corresponding energy consumption, respectively. Experimental results are shown in Figure \ref{fig:7}. It is worth noting that we only compare ADUMBO with USeMO, which is proven to be the most close baseline algorithm with ADUMBO as shown in above experimental results. Obviously, the Pareto front found by ADUMBO always dominates USeMO under different dataset types and sizes. 
For instance, considering the high performance requirement, ADUMBO is able to achieve a better model training time while reducing the total energy consumption by 9.41\%, 6.35\% and 3.68\%  compred with USeMO while achieving a better model training time under LeNet-MNIST, LeNet-EMNIST-50\% and LeNet-EMNIST, respectively. On the other hand, considering the energy conservation requirement, ADUMBO can also always further reduce the energy consumption with less associated model training time. 

Therefore, the proposed HyperTuner framework can indeed well adapt to different datasets.

\subsubsection{\textbf{Different optimization objectives}} 
\begin{figure*}
 \centering
 \subfigure[]{    
  \label{fig:8a}\includegraphics[width=0.32\textwidth]{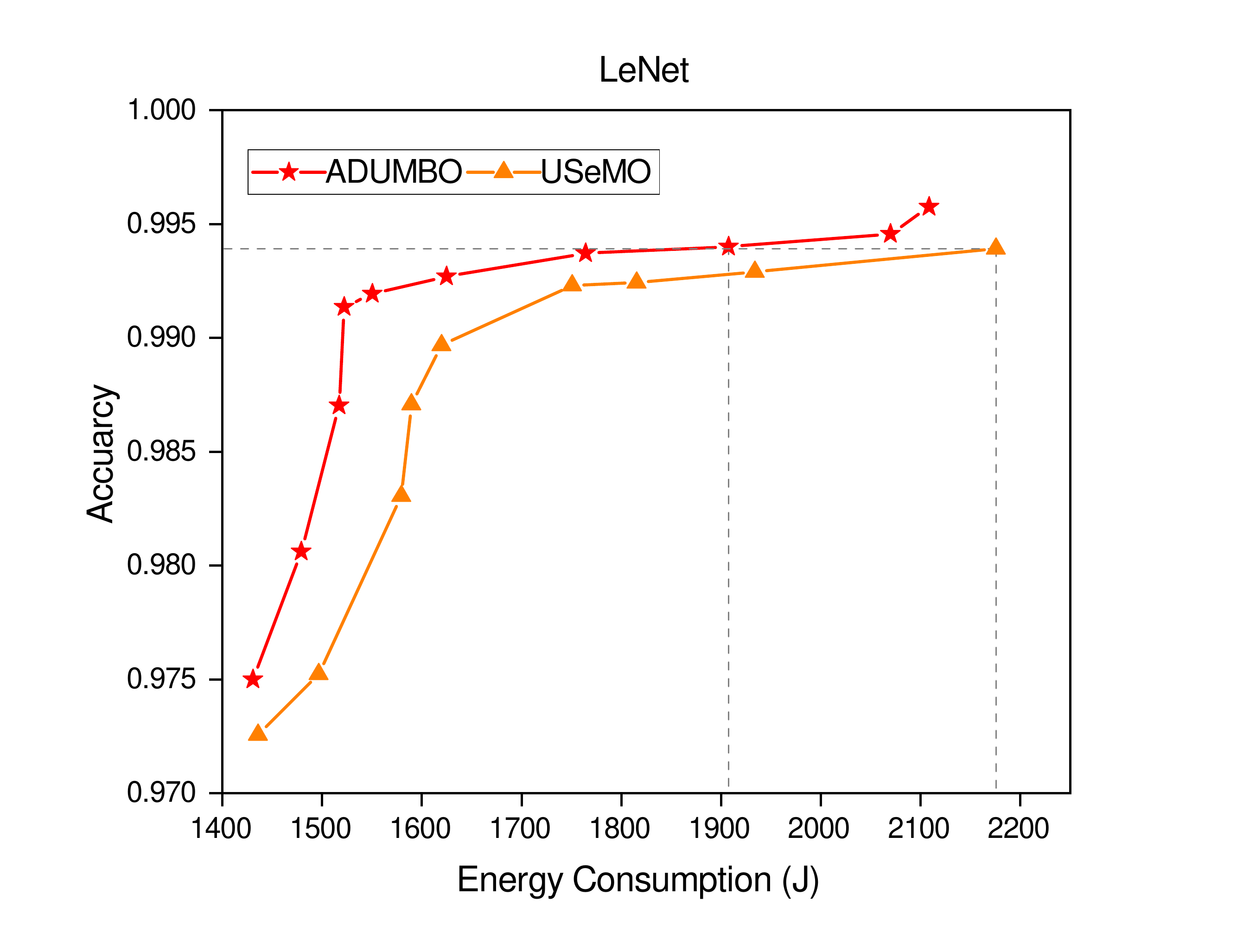}}
   \subfigure[]{
    \label{fig:8b}\includegraphics[width=0.32\textwidth]{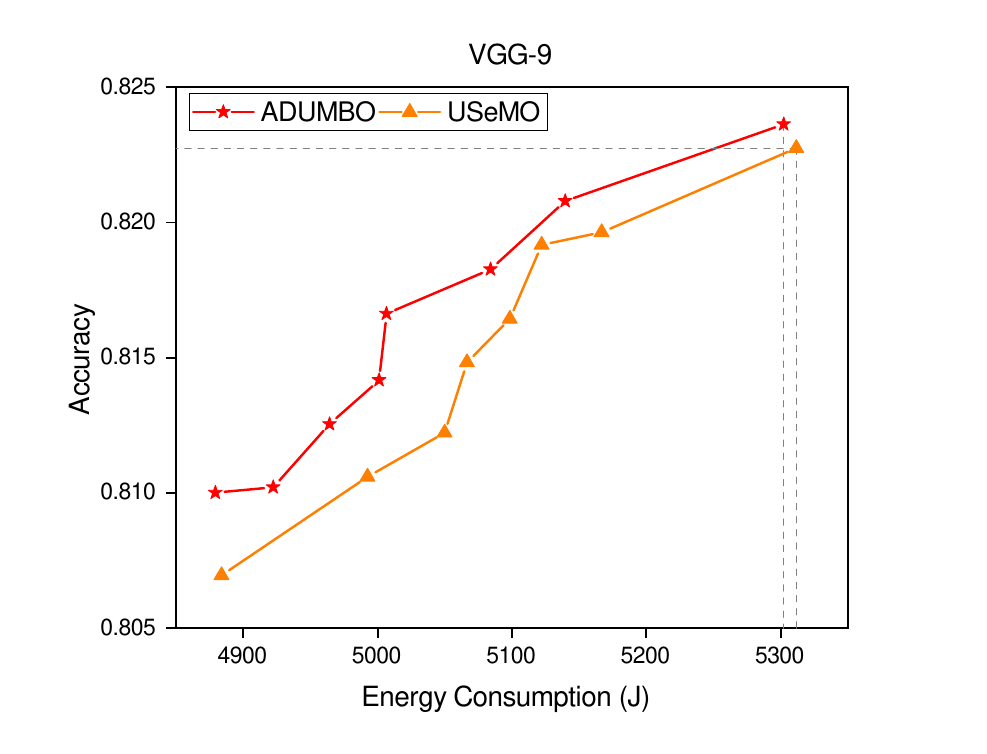}}
   \subfigure[]{    
  \label{fig:8c}\includegraphics[width=0.32\textwidth]{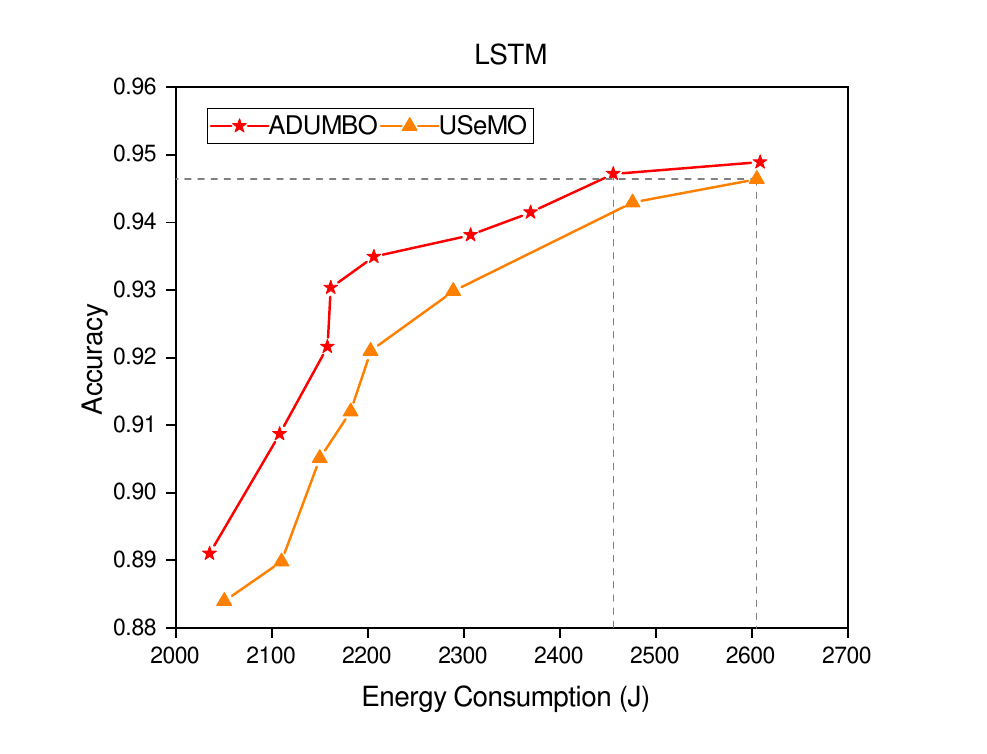}}
 \caption{The adaptability of HyperTuner to different optimization objectives.}
 \label{fig:8}
\end{figure*}

In practice, there is rarely a clear-cut single optimization objective for data analytic services. Instead, the multiple optimization intentions including model training time, energy consumption and model accuracy are of equal importance for service providers. 
Previous experiments in Sec. \ref{sec:6b} have verified the performance of HyperTuner in simultaneously optimizing the model training time and energy consumption. To further evaluate the adaptability of HyperTuner for other optimization intentions, we inherited the previous settings of model training tasks and changed the two optimization objectives to the model accuracy and energy consumption. Experimental results are shown in Figure \ref{fig:8}. Compared with the most close algorithm USeMO, ADUMBO is always  able to find a better Pareto-optimal configuration set under all the three DNN models. Specifically, for the high accuracy requirement, to achieve a model accuracy not-lower-than the best of USeMO, ADUMBO can save the associated energy consumption by as much as 12.32\%, 0.75\% and 5.76\% under LeNet, VGG-9 and LSTM, respectively. Therefore, the proposed cross-layer multi-objective hyperparameter auto-tuning framework HyperTuner can well adapt to different optimization objectives for data analytic services.  

\subsubsection{\textbf{Different machine learning frameworks}} Previous experiments are all conducted with a distributed TensorFlow cluster executing three deep learning model training tasks. In practice, however, there are actually quite a few other frameworks such as PyTorch, BigDL and Spark can be chosen. To evaluate the adaptability of HyperTuner to other machine learning frameworks as well as model training tasks, we extend it to support configuration auto-tuning for a Spark-based data analytic cluster and utilize the Spark MLlib library to implement three new model training tasks namely logistic regression (LR), decision tree classifier (DT) and support vector machine (SVM). After that, We compare ADUMBO with four baseline algorithms including Random search, PABO, FlexiBO and USeMO regarding the model accuracy and energy consumption as two optimization objectives. The number of total configuration evaluation times are all set to be 70. In addition, to shrink the high-dimensional candidate configuration space, MOPIR method is utilized to select 15 critical parameters for each model training task. We list the chosen parameters of each layer in Tab. \ref{tab5}. 

\begin{figure*}
 \centering
 \subfigure[]{    
  \label{fig:9a}\includegraphics[width=0.32\textwidth]{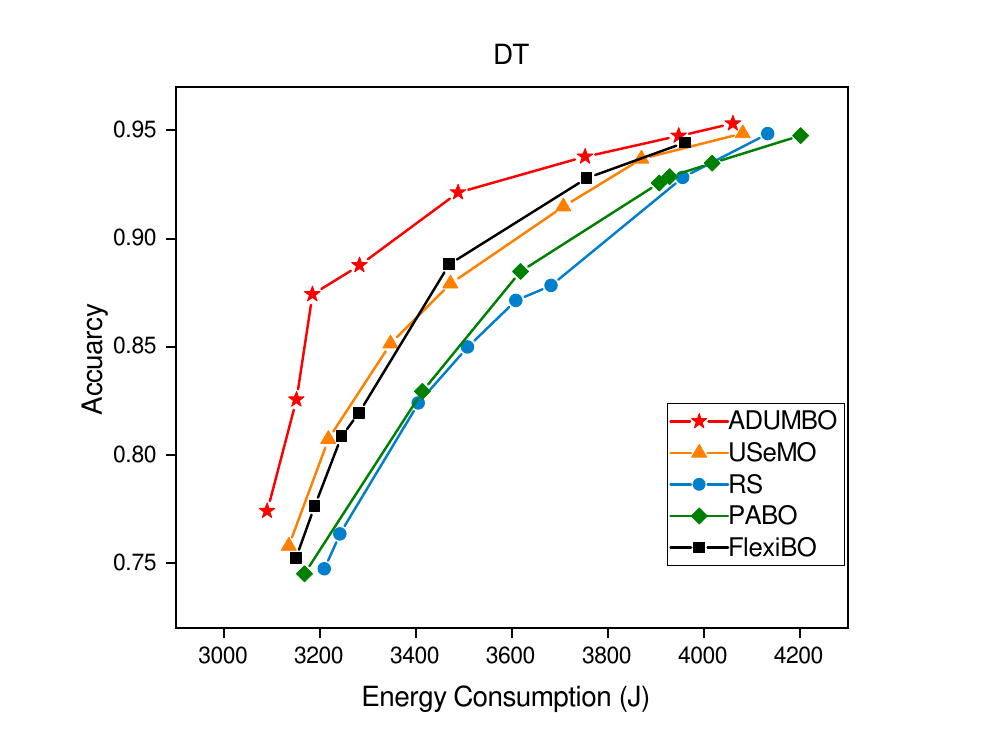}}
   \subfigure[]{
    \label{fig:9b}\includegraphics[width=0.32\textwidth]{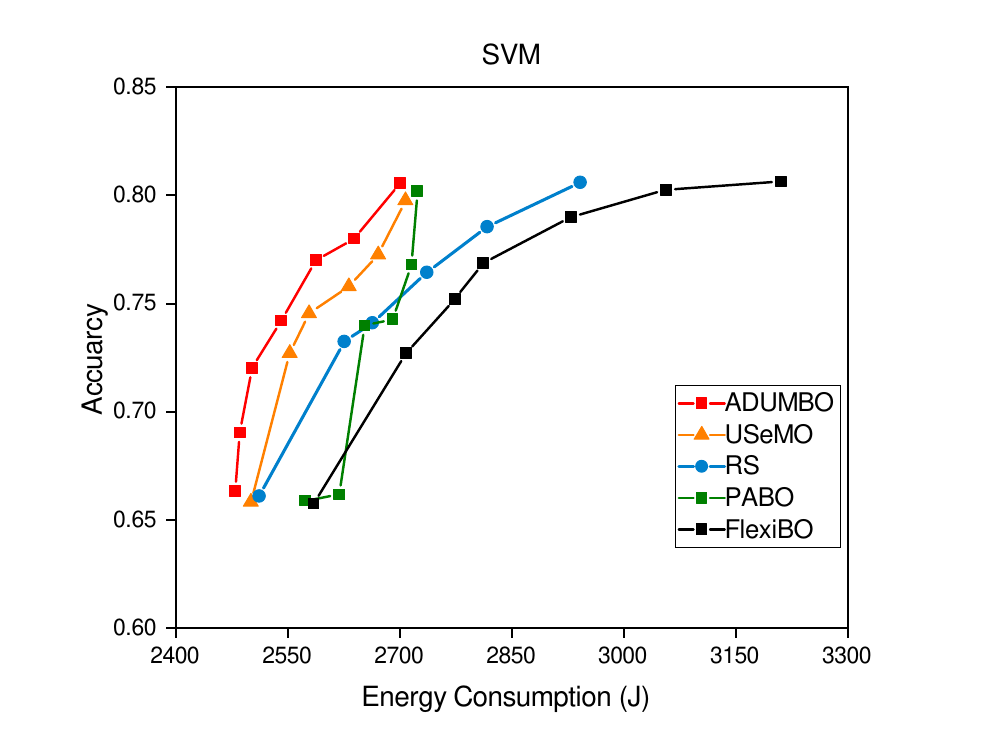}}
   \subfigure[]{    
  \label{fig:9c}\includegraphics[width=0.32\textwidth]{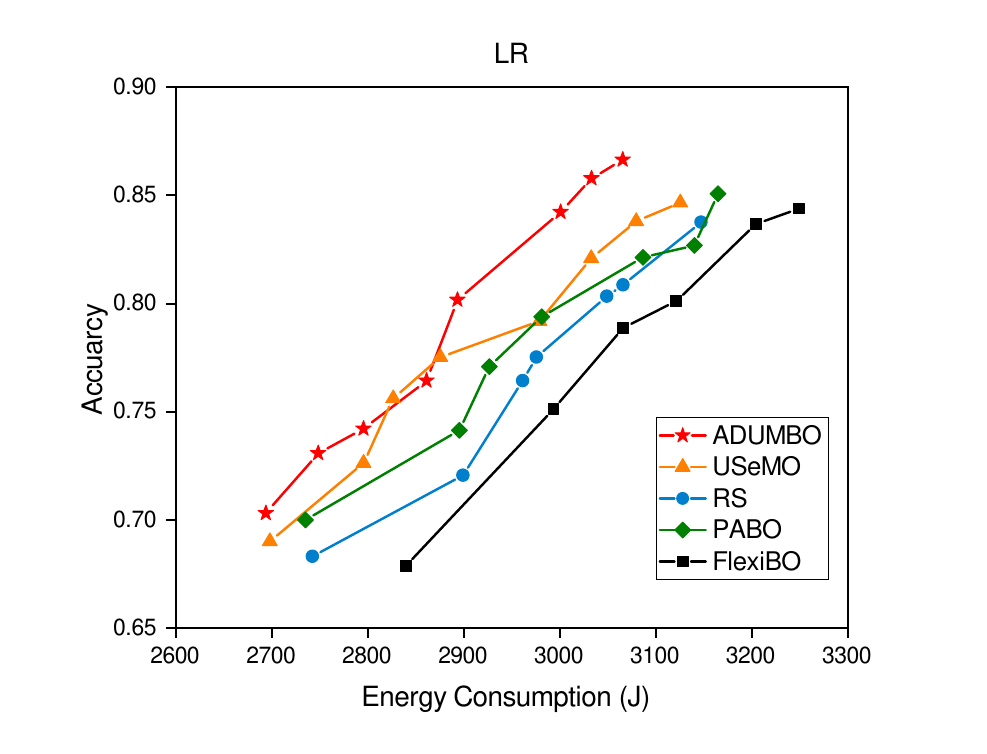}}
  \subfigure[]{ 
    \label{fig:9d}\includegraphics[width=0.32\textwidth]{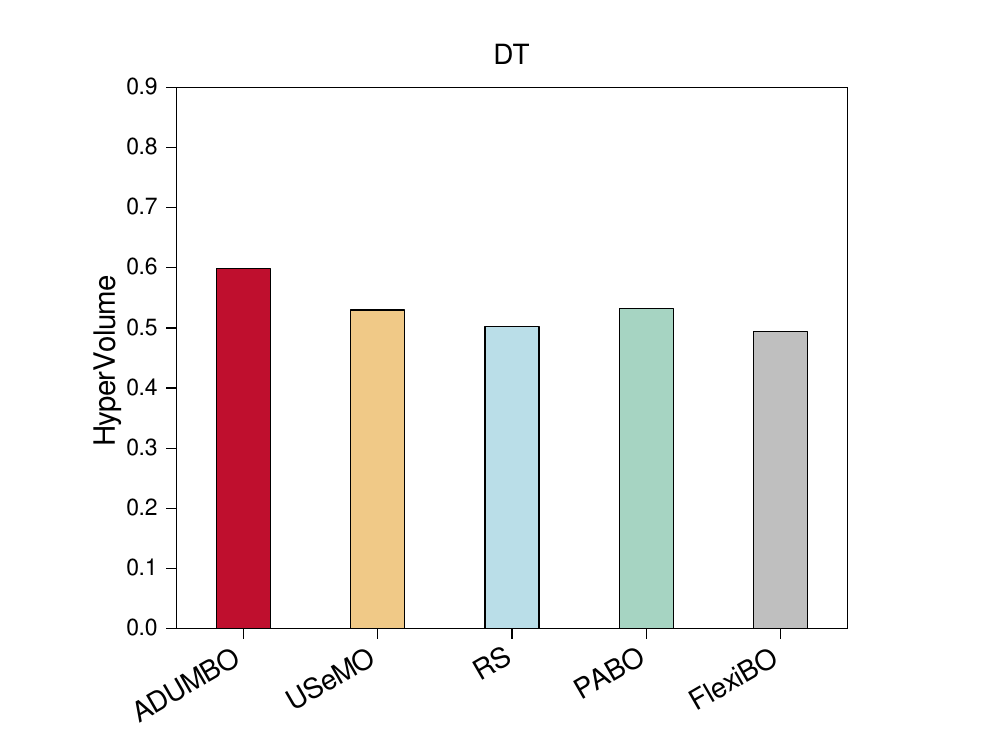}}
   \subfigure[]{
    \label{fig:9e}\includegraphics[width=0.32\textwidth]{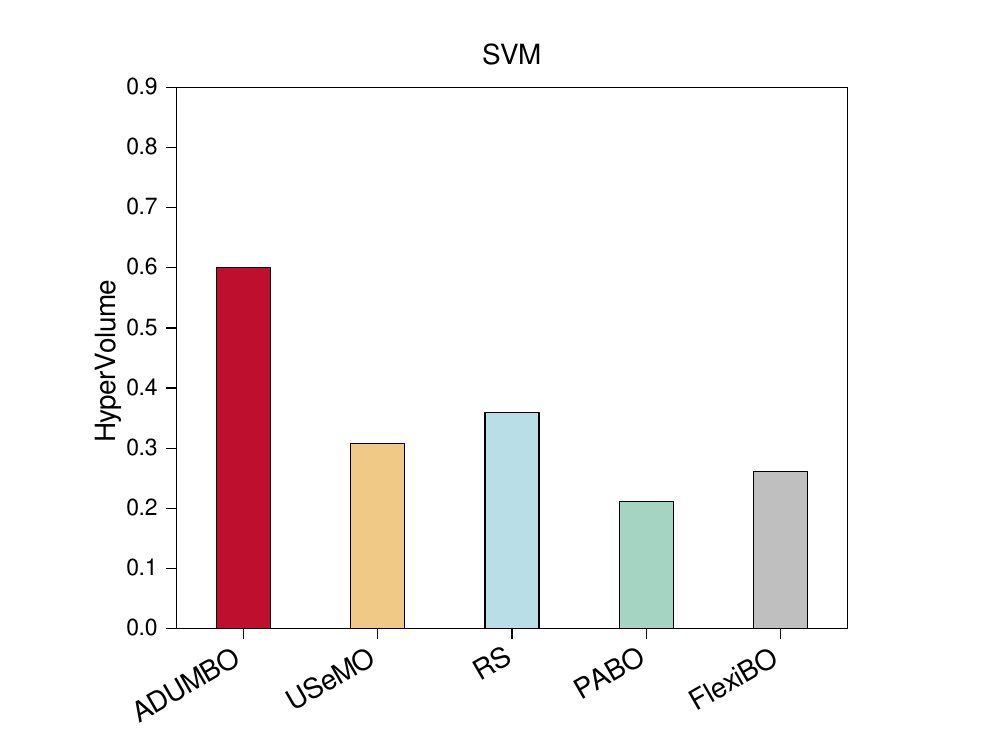}}
   \subfigure[]{    
  \label{fig:9f}\includegraphics[width=0.32\textwidth]{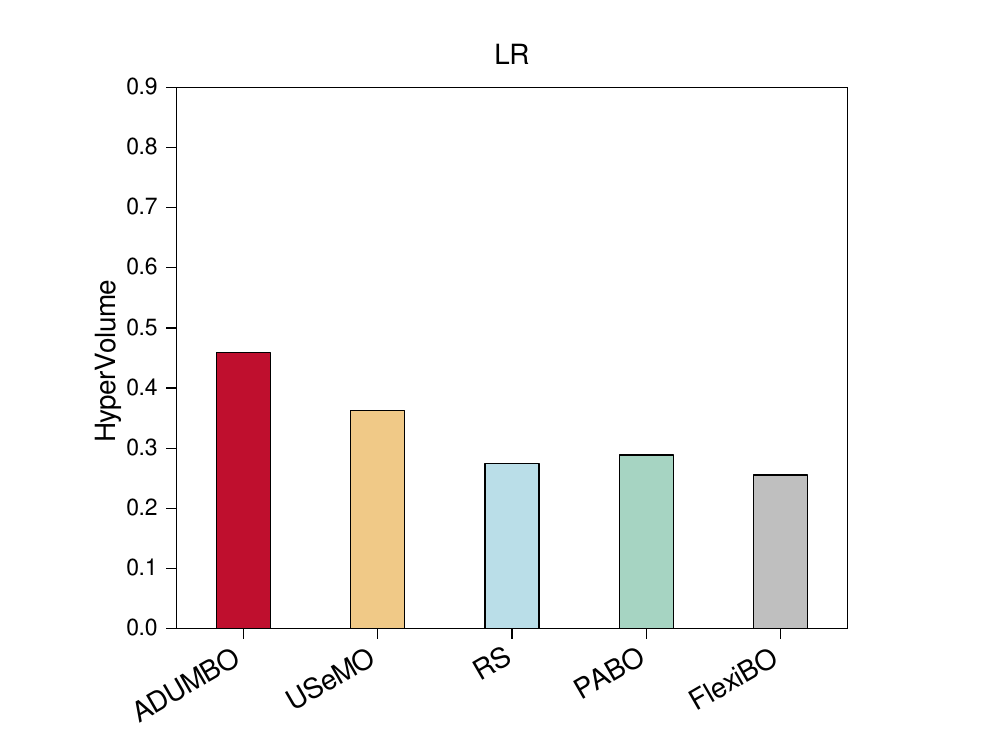}}
 \caption{The adaptability of HyperTuner to a Spark-based machine learning cluster.}
 \label{fig:9}
\end{figure*}

Figure \ref{fig:9a}-\ref{fig:9c} show the Pareto-optimal configuration set found by ADUMBO and the four baseline algorithms. It is obvious that ADUMBO is always able to find a better Pareto front for the three different machine learning model training tasks and the vast majority Pareto configurations found by ADUMBO are non-dominant by any other baselines. For example, under the high accuracy optimization requirement, ADUMBO can always find a dominant configuration  with the highest model accuracy and least energy consumption. Taking the LR model training task as an instance, ADUMBO can achieve a no-lower-than model accuracy while consuming respectively 4.90\%, 4.31\%, 7.09\% and 3.06\% less energy compared with Random search, PABO, FlexiBO and USeMO. To further give a deep quantitative analysis, we again leverage the HyperVolume indicator to measure the quality of the achieved Pareto fronts by different algorithms. As shown in Figure \ref{fig:9d}-\ref{fig:9f}, the Pareto-optimal set of ADUMBO is indeed always able to achieve a higher HyperVolume value under the 3 different model training tasks. That is to say, the Pareto front found by ADUMBO owns a higher
convergence (minimizing the gap between candidate solutions and the true Pareto-optimal set) and diversity (maximizing the distribution of candidate solutions)  than all the other baselines. These experimental results verify that HyperTuner can indeed well adapt to different machine learning frameworks as well as different data analytic services.
\begin{table*}[htbp]
\centering
\caption{The cross-layer critical parameters for SVM and Logisitic Regression model training tasks.}
\begin{center}
\scalebox{0.8}{
\begin{tabular}{|c|c|l|}
\hline
\textbf{Parameter} &\textbf{Range} &\multicolumn{1}{c|}{\textbf{Description}} \\
\hline 
\textit{numIterations}& [10, 100, 1000]& The number of iterations to run \\
 \cline{1-3}  
 \textit{stepSize} &[0.1, 1, 10, 100]& The step size for gradient descent \\
 \cline{1-3}  
 \textit{miniBatchFraction}&	[0.01, 0.1, 0.5, 1]& \makecell[l]{The fraction of the total data that is sampled in each iteration,\\ to compute the gradient direction}	\\
 \cline{1-3}   
  \textit{regParam}&	[0.01, 0.1, 1, 10] & Regularization parameter	 \\
  \cline{1-3}  
\textit{spark.default.parallelism}& 8-128&\makecell[l]{The largest number of partitions in a parent RDD for distributed\\ shuffle operations}\\
 \cline{1-3}  
 \textit{ spark.executor.memory}&	1024-4028	& \makecell[l]{The amount of memory owned by the executor process, in MB} \\
 \cline{1-3}   
\textit{ spark.io.compression.blockSize}&2-128	&\makecell[l]{Block size used in LZ4(Snappy) compression, in KB} \\
 \cline{1-3} 
 \textit{spark.speculation.interval} &10-1000& \makecell[l]{How often Spark will check for tasks to speculate, in millisecond }\\
 \cline{1-3} 
  \textit{spark.speculation.quantile}&0.3-0.9	& \makecell[l]{Fraction of tasks must be complete before speculation is enabled\\for a particular stage} \\
  \hline
\textit{swappiness}&0-100&Frequency to drop caches and swap out application pages\\
 \cline{1-3}
\textit{dirty\_expire\_centisecs}&	100-10000& Max duration interval of dirty pages\\
 \cline{1-3}
\textit{nr\_requests}&	64-256	&number of requests allocated in block layer\\
 \cline{1-3}
\textit{read\_ahead\_kb}&	64-512	& How much extra data kernel reads from disk\\
\cline{1-3}
\textit{cpu\_ferq}	&\makecell{[800MHZ, 900MHz, 1.1GHz, \\ 1.2GHz, 1.4GHz, 1.5 GHz, 1.7GHz,\\1.8GHz, 2GHz, 2.2GHz, 2.4GHz,\\ 2.6GHz, 2.7 GHz, 2.9 GHz]} &\makecell[l]{Available frequency provided by the \textit{userspace} governor}\\
\hline
\end{tabular}
}
\label{tab5}
\end{center}
\end{table*}
\begin{table*}[htbp]
\centering
\caption{The cross-layer critical parameters for Decision Tree model training tasks.}
\begin{center}
\scalebox{0.8}{
\begin{tabular}{|c|c|l|}
\hline
\textbf{Parameter} &\textbf{Range} &\multicolumn{1}{c|}{\textbf{Description}} \\
\hline 
 \cline{1-3}  
\textit{impurity}&['gini', 'entropy']&The measure used to choose between candidate splits \\
 \cline{1-3}  
\textit{maxDepth}&[3,5,10,20,25, 30] &The maximum depth of the decision tree\\
 \cline{1-3}  
\textit{maxBins}&[5,10,50,100,200]& The number of bins used when discretizing continuous feature \\
\cline{1-3}  
\textit{spark.default.parallelism}& 8-128&\makecell[l]{The largest number of partitions in a parent RDD for distributed\\ shuffle operations}\\
 \cline{1-3}  
 \textit{ spark.executor.memory}&	1024-4028	& \makecell[l]{The amount of memory owned by the executor process, in MB} \\
 \cline{1-3}   
\textit{ spark.io.compression.blockSize}&2-128	&\makecell[l]{Block size used in LZ4(Snappy) compression, in KB} \\
 \cline{1-3} 
 \textit{spark.speculation.interval} &10-1000& \makecell[l]{How often Spark will check for tasks to speculate, in millisecond }\\
 \cline{1-3} 
  \textit{spark.speculation.quantile}&0.3-0.9	& \makecell[l]{Fraction of tasks must be complete before speculation is enabled\\for a particular stage} \\
 \cline{1-3} 
\textit{spark.driver.memory}&1024-4028 &\makecell[l]{The amount of memory owned by the diver process, in MB}\\
 \cline{1-3} 
\textit{spark.kryoserializer.buffer}&2-128 &Initial size of Kryo’s serialization buffer, in KB\\
\hline
\textit{swappiness}&0-100&Frequency to drop caches and swap out application pages\\
 \cline{1-3}
\textit{dirty\_expire\_centisecs}&	100-10000& Max duration interval of dirty pages\\
 \cline{1-3}
\textit{nr\_requests}&	64-256	&number of requests allocated in block layer\\
 \cline{1-3}
\textit{read\_ahead\_kb}&	64-512	& How much extra data kernel reads from disk\\
\cline{1-3}
\textit{cpu\_ferq}	&\makecell{[800MHZ, 900MHz, 1.1GHz, \\ 1.2GHz, 1.4GHz, 1.5 GHz, 1.7GHz,\\1.8GHz, 2GHz, 2.2GHz, 2.4GHz,\\ 2.6GHz, 2.7 GHz, 2.9 GHz]} &\makecell[l]{Available frequency provided by the \textit{userspace} governor}\\
\hline
\end{tabular}
}
\label{tab6}
\end{center}
\end{table*}

\section{Related Work}
\label{sec:7}
In this section, we first describe several categories of multi-objective hyperparameter optimization algorithm and explain why we design and implement a BO-based algorithm in HyperTuner. After that, we also introduce several representative cross-layer hyperparameter optimization methods and describe the differences between HyperTuner and these recent work. 
\subsection{Naive multi-objective HPO algorithms}
\label{section:7a}
Grid Search\cite{montgomery2017design}, Random Search\cite{bergstra2012random} and Scalarization\cite{scalarization} are three most representative Naive approaches for multi-objective HPO. Specifically, Grid Search and Random Search are two algorithms original from single-objective HPO. Since the multiple objectives further expand the candidate configuration space and configuration evaluation is an expensive job in practice, both methods fail to tackle with the rigorous challenge of high sample-efficiency. The core idea of scalarization-based methods is to create several scalar single-objective optimization problems in place of the original multi-objective one. However, the interaction between multiple objectives is agnostic, and it is difficult to set appropriate scaling parameters without prior experience or expertise. 
\subsection{Multi-objective evolutionary algorithms}
Evolutionary algorithms are typical black-box optimization algorithms inspired by principles of natural evolution, which can be used to solve the multiple-objective hyperparameter optimization problem. The non-dominated sorting genetic algorithm (NSGA-II) \cite{nsga} is one of the most popular MOEAs. It uses specific types of cross and mutation operations to generate genetically line offspring, then selects the next generation based on non-dominant sorting and crowding distance comparison.
However, since the non-dominated sorting becomes less discriminative and the left and right neighbor are often different in each objective, it may lose attraction along with the number of total objectives.  Different from NSGA-II, the Multi-Objective Evolutionary Algorithm based on Decomposition (MOEA/D)\cite{moea} 
 decomposes the multi-objective problem into a finite number of single scalar optimization problems that are optimized simultaneously. Thus, it is able to scale well with the number of objectives. However, it is challenging to identify scalarization methods without a good knowledge of the Pareto frontier. In addition, another common disadvantage of evolutionary algorithms is the relatively slow convergence speed. As a result, the associated huge number of configuration evaluation prevents these methods from being applied into configuration auto-tuning tasks for data analytic services in practice.

\subsection{Multi-objective Bayesian optimization algorithms}
Bayesian optimization algorithm have already demonstrated its sample-efficiency in solving single-objective black-box optimization problems, especially the ones with high evaluation cost such as hyperparameter optimization for data analytic services. There are two typical ways to extend vanilla BO to multi-objective BO. First of all, methods such as ParEGO \cite{knowles2006parego} directly scale the multiple objectives into a single objective in a specific manner and turn the original multi-objective optimization problem into a single-objective one. Although simple and fast, a common drawback of these methods lies in the difficulty to determine the appropriate scalarization manner to avoid falling into sub-optimal solutions which may be far from the practical Pareto front. Second, methods in \cite{ehi}\cite{sms-ego}\cite{hernandez2016predictive}\cite{ belakaria2020uncertainty} choose to build a surrogate model respectively for each objective and solve a tailored optimization problem based on acquisition functions to recommend promising solutions. For instance, EHI\cite{ehi} individually updates different surrogate models for each objective and extends the standard acquisition functions to measure the expected improvement in Pareto hypervolume (PHV) metric to generate configuration for next iteration. SMS-EGO  \cite{sms-ego} expands the idea of EHI via a novel padding criterion. Unfortunately, finding a configuration with optimal PHV values based on the acquisition functions needs a large amount of calculation and thus scales very poorly with the number of objectives. On the other hand, PESMO\cite{hernandez2016predictive} proposes entropy-based acquisition function to measure the information gained about the true Pareto-optimal set when determining the next promising solution. However, it is also computationally expensive to optimize the acquisition function employed in PESMO. To mitigate these computational costs, UseMO \cite{belakaria2020uncertainty} first establishes a separate surrogate model for each objective, and then the acquisition function of multiple models is jointly optimized as a cheap MOO problem to identify the most promising candidates and pick the best candidate based on the uncertainty metric. Unfortunately, this uncertainty metric only focuses on exploration the variances of current surrogate models and fails to help achieve the exploitation-exploration balance with the increasing iteration times. Inspired by USeMO, the proposed ADUMBO algorithm in HyperTuner selects the most promising candidate configuration from the generated Pareto-optimal set via maximizing a new well-designed metric, which can adaptively leverage the uncertainty across all the surrogate
models along with the iteration times. Experimental results in this paper evaluate the advantages of ADUMBO over prior multi-objective BO algorithms. 

\subsection{Cross-layer hyperparameter optimization for data analytic services}
In practice, the performance of the machine learning models has an inseparable relationship with their underlying deployed platform. As a result, hardware and system parameters are also critical factors should be considered when conducting model hyperparameter optimization. Therefore, in the past few years, some advanced methods\cite{zela,hardware,bayestuner} have tried to automatically optimize the model training and inference tasks in the joint space of network architecture, hyperparameters and computer system stack for a certain single objective. For example, BayesTuner\cite{bayestuner} leverages a constrained Bayesian optimization to find the optimal hyperparameter and hardware configuration (e.g., batch size, number of CPU cores, allocated memory) to reduce inference energy consumption under performance constraints. Recently, some studies\cite{lokhmotov2018multi, iqbal2020flexibo} begin to focus on the cross-layer multi-objective hyperparameter optimization problem. 
Specifically, Lokhmotov et al.\cite{lokhmotov2018multi} use a Collective Knowledge workflow to study the execution time and accuracy trade-offs for MobileNets on two specified development platforms. However, this work lacks adaptability to other practical data analytic scenarios. 
As the most related work, FlexiBO \cite{iqbal2020flexibo} is motivated by a multi-objective Bayesian optimization algorithm called Pareto Active Learning (PAL)\cite{pal}. It uses the same Pareto front construction method with PAL and introduces a cost-aware acquisition function to sample the next configuration for evaluation. However, its effectiveness and efficiency highly depends on the assumption of a discrete candidate configuration space since it has to traverse the whole space to generate the next sample configuration. Therefore, in order to address the challenging high-dimensional black-box multi-objective optimization problem, HyperTuner first leverages MOPIR to select critical parameters and then utilizes ADUMBO to efficiently explore the shrinked candidate configuration space to find Pareto-optimal configuration set for the specified multiple objectives. 

\section{Conclusion and Future Work}
\label{sec:8}
In this paper, we propose HyperTuner, a cross-layer multi-objective hyperparameter auto-tuning framework for data analytic services. To address the formulated high-dimensional black-box multi-objective optimization problem, HyperTuner first conduct multi-objective parameter importance ranking with its MOPIR algorithm and
then leverage the proposed ADUMBO algorithm to find the Pareto-optimal configuration set. We evaluate HyperTuner on our local distributed TensorFlow cluster with 3 DNN model training tasks. Experimental results show that compared with other four baseline multi-objective black-box optimization algorithms, HyperTuner is always able to find a better Pareto-optimal configuration set superior in both convergence and diversity. In addition, experiments with different training datasets, different optimization objectives and different machine learning platforms verify that HyperTuner can well adapt to the various application scenarios of data analytic services. 

In the future, we will add support in HyperTuner for GPU-based machine learning systems since the performance as well as power consumption of GPU devices draw more and more attention from data analytic service providers. In addition, involving cloud resource configuration into the hyperparameter optimization problem is also a promising and interesting issue, which is our ongoing work based on HyperTuner.

\begin{acknowledgement}
The authors would like to thank the anonymous reviewers for their valuable comments and suggestions. This work was supported by the National Natural Science Foundation of China under Grant
61902440, 62272001 and U1811462.
\end{acknowledgement}

\bibliographystyle{fcs}
\bibliography{ref}

\begin{wrapfigure}{l}{25mm}
    \includegraphics[width=1in,height=1.25in,clip]{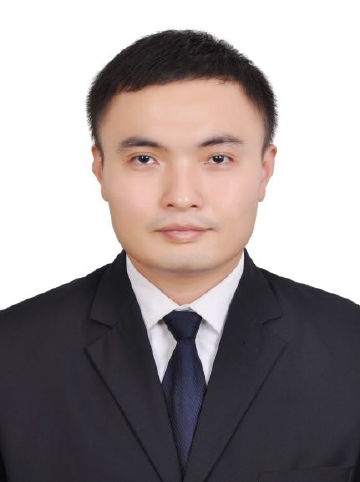}
\end{wrapfigure}\par
\noindent{\footnotesize \textbf{Hui Dou} is currently an associated professor in School of Computer Science and Technology at Anhui University, master tutor. Dr. Dou graduated from the department of computer science of Xi’an Jiaotong University with a Ph.D. degree in 2017. After that, he worked as a senior engineer in Huawei from Sep, 2017 to Jun, 2018. From Aug, 2018 to Aug, 2020, he undertook post-doctoral research with associated professor Pengfei Chen at Sun Yat-sen University. Dr. Dou is now interested in AI-driven performance and energy consumption optimization for distributed systems, including configuration tuning and workload scheduling. Until now, Dr. Dou has published more than 10 papers in some international conferences and journals such as ICDCS, ICPP, IEEE TSUSC and FCS.}\par
  
\begin{wrapfigure}[8]{l}{25mm} 
    \includegraphics[width=1in,height=1.25in,clip]{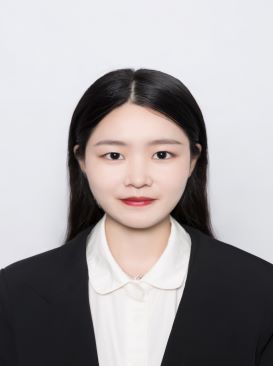}
\end{wrapfigure}\par
\noindent{\footnotesize \textbf{Shanshan Zhu} received her BS in computer science and technology from Anhui Jianzhu University  in 2021, and now is a master student in the School of Computer Science and Technology at Anhui University. Her current research topics mainly focus on hyperparameter optimization for deep learning models.}\par

\begin{wrapfigure}{l}{25mm} 
    \includegraphics[width=1in,height=1.25in,clip]{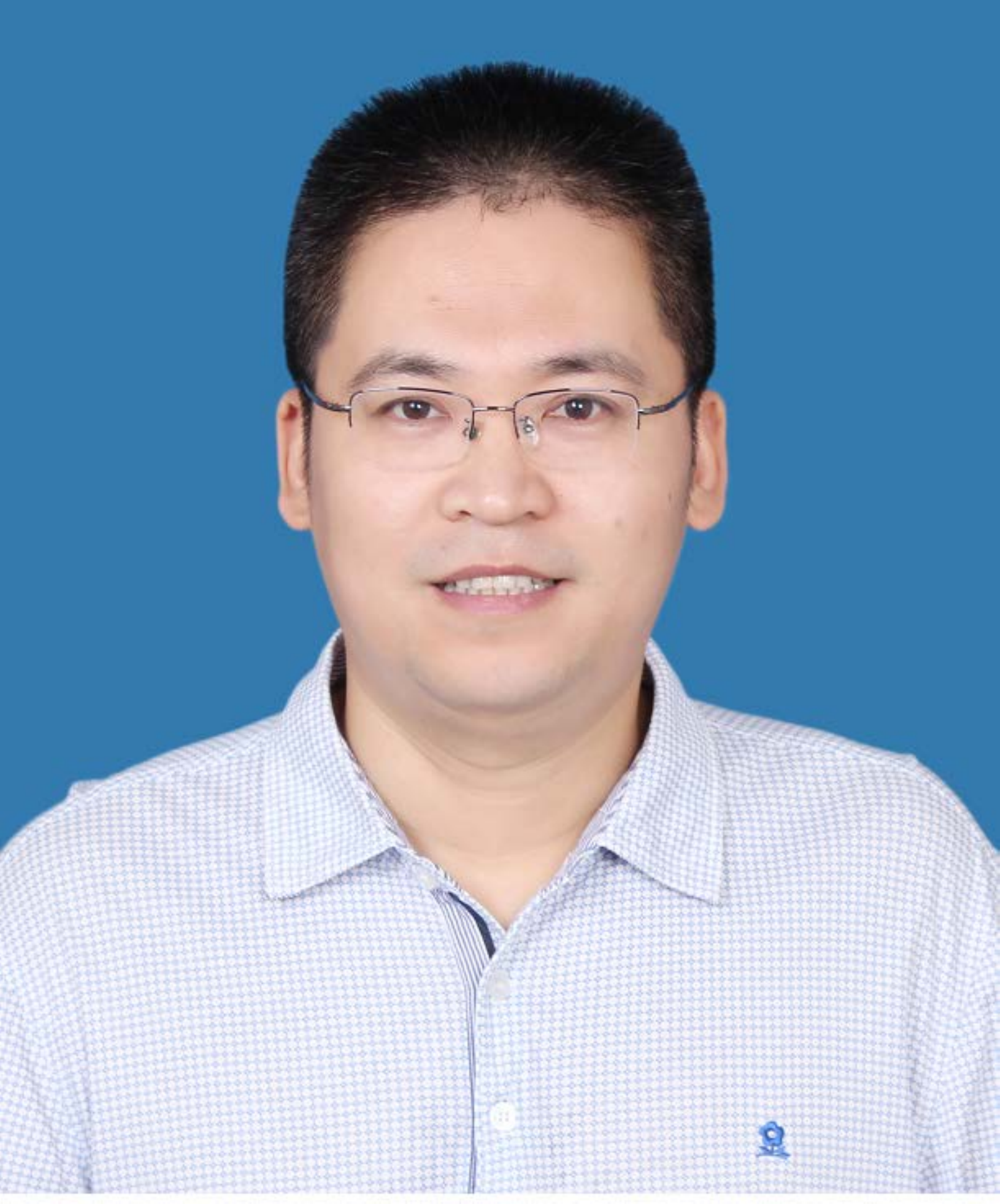}
\end{wrapfigure}\par
\noindent{\footnotesize \textbf{Yiwen Zhang} received the Ph.D. degree in management science and engineering from the Hefei University of Technology, in 2013. He is currently a Professor with the School of Computer Science and Technology, Anhui University. Meanwhile, he is a Ph.D. advisor. His research interests include service computing, cloud computing, and big data analytics. Until now, Dr. Zhang has published more than 70 papers in some international conferences including IEEE ICSOC, IEEE ICWS and journals including IEEE TSC, TSMC, TMC.}\par

\begin{wrapfigure}{l}{25mm} 
    \includegraphics[width=1in,height=1.25in,clip]{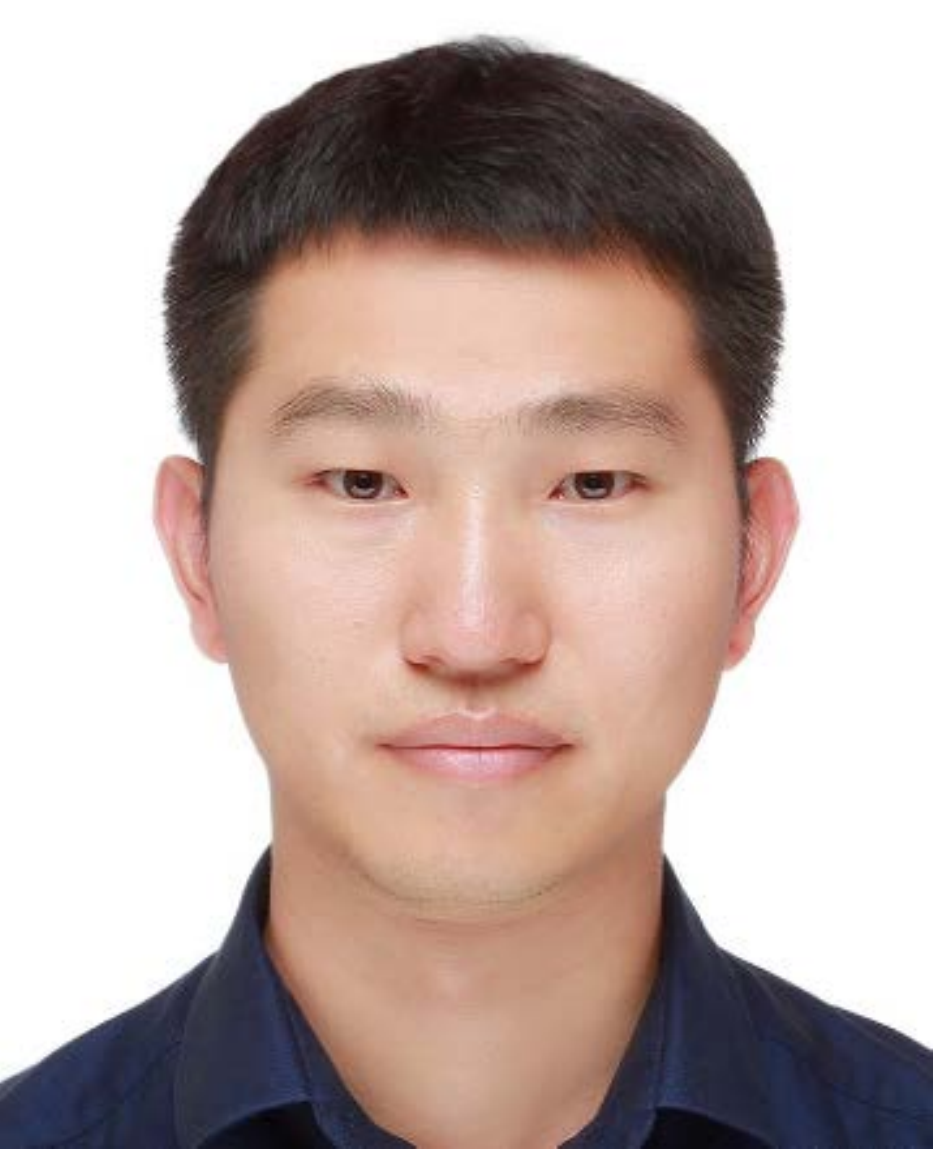}
\end{wrapfigure}\par
\noindent{\footnotesize \textbf{Pengfei Chen} is currently an associated professor in School of Computer Science and Engineering at Sun Yat-sen University. Meanwhile, he is a Ph.D. advisor. Dr. Chen graduated from the department of computer science of Xi’an Jiaotong University with a Ph.D. degree in 2016. After graduation, Dr. Chen worked as a Research Scientist in IBM Research China during Jun, 2016-Jan, 2018. Now, he is interested in distributed systems, AIOps, cloud computing, Microservice and BlockChain. Especially, he has strong skills in cloud computing. Until now, Dr. Chen has published more than 30 papers in some international conferences including IEEE INFOCOM, WWW, IEEE ICSOC, IEEE ICWS, IEEE BigData and journals including IEEE TDSC, IEEE TR, IEEE TSC, IEEE TETC, IEEE TCC.}\par
\begin{wrapfigure}{l}{25mm} 
    \includegraphics[width=1in,height=1.25in,clip]{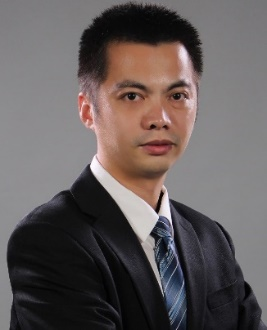}
\end{wrapfigure}\par
\noindent{\footnotesize \textbf{Zibin Zheng} is currently a Professor at School of Computer Science and Engineering with Sun Yat-sen University. Zibin Zheng received his Ph.D. degree from Chinese University of Hong Kong, in 2011. His research interests include blockchain, services computing, software engineering, and financial big data. He published over 120 international journal and conference papers, including 3 ESI highly cited papers. According to Google Scholar, his papers have more than 7000 citations, with an H-index of 42. Prof. Zheng was a recipient of several awards, including the Top 50 Influential Papers in Blockchain of 2018 and the ACM SIGSOFT Distinguished Paper Award at ICSE 2010.}\par

\end{sloppypar}
\end{document}